\documentclass{article}

\usepackage{microtype}
\usepackage{graphicx}
\usepackage{subfigure}
\usepackage{booktabs} 

\usepackage{hyperref}
\usepackage{multirow}

\usepackage[accepted]{icml2024}

\usepackage{algorithm}
\usepackage{algorithmic}
\usepackage{newfloat}
\usepackage{listings}
\usepackage{graphicx}
\graphicspath{ {images/} }
\usepackage{caption}
\usepackage{subfigure}
\usepackage{subcaption}
\usepackage{nicefrac}
\usepackage{amssymb}
\usepackage{amsthm}
\usepackage{newtxmath}
\usepackage{scrextend}
\usepackage{numprint}
\usepackage{makecell}
\usepackage{pbox}

\newtheorem{theorem}{Theorem}[section]

\newtheorem{lemma}[theorem]{Lemma}
\newtheorem*{lemma*}{Lemma}

\npthousandsep{,}

\usepackage[utf8]{inputenc} 
\usepackage[T1]{fontenc}    
\usepackage{url}            
\usepackage{booktabs}       
\usepackage{amsfonts}       
\usepackage{microtype}      
\usepackage{xcolor}     

\newcommand{\x}{\mathbf{x}}
\newcommand{\e}{\mathbf{e}}

\newcommand{\w}{\mathbf{w}}

\newcommand{\h}{\mathbf{h}}

\newcommand{\amirg}[1]{\textcolor{red}{AG:#1}}
\newcommand{\maya}[1]{\textcolor{blue}{M:#1}}

\newcommand{\reals}{\mathbb{R}}
\newcommand{\struct}{graph-structure}
\newcommand{\structs}{graph-structures}

\icmltitlerunning{Graph Neural Networks Use Graphs When They Shouldn't}

\begin{document}

\twocolumn[
\icmltitle{Graph Neural Networks Use Graphs When They Shouldn't}



\icmlsetsymbol{equal}{*}

\begin{icmlauthorlist}
\icmlauthor{Maya Bechler-Speicher}{1}
\icmlauthor{Ido Amos}{2}
\icmlauthor{Ran Gilad-Bachrach}{3}
\icmlauthor{Amir Globerson}{1}

\end{icmlauthorlist}

\icmlaffiliation{1}{Blavatnik School of Computer Science,
  Tel-Aviv University}
\icmlaffiliation{2}{School of Electrical Engineering,
  Tel-Aviv University}
\icmlaffiliation{3}{Department of Bio-Medical Engineering and Edmond J. Safra Center for Bioinformatics, Tel-Aviv University}

\icmlcorrespondingauthor{Maya Bechler-Speicher}{mayab4@mail.tau.ac.il}

\icmlkeywords{Graph Neural Networks, Graph Learning}

\vskip 0.3in
]



\printAffiliationsAndNotice{}  

\begin{abstract}
Predictions over graphs play a crucial role in various domains, including social networks and medicine.
Graph Neural Networks (GNNs) have emerged as the dominant approach for learning on graph data.
Although a graph-structure is provided as input to the GNN, in some cases the best solution can be obtained by ignoring it.
While GNNs have the ability to ignore the graph-structure in such cases, it is not clear that they will.
In this work, we show that GNNs actually tend to overfit the given graph-structure. Namely, they use it even when a better solution can be obtained by ignoring it.
We analyze the implicit bias of gradient-descent learning of GNNs and prove that when the ground truth function does not use the graphs, GNNs are not guaranteed to learn a solution that ignores the graph, even with infinite data.
We examine this phenomenon with respect to different graph distributions and find that regular graphs are more robust to this overfitting.  
We also prove that within the family of regular graphs, GNNs are guaranteed to extrapolate when learning with gradient descent.
Finally, based on our empirical and theoretical findings, we demonstrate on real-data how regular graphs can be leveraged to reduce graph overfitting and enhance performance.
\end{abstract}
\section{Introduction}


Graph labeling problems arise in many domains, from social networks to molecular biology.
In these settings, the goal is to label a graph or its nodes given information about the graph. The information for each graph instance is typically provided in the form of the graph-structure (i.e., its adjacency matrix) as well as the features of its nodes. 

Graph Neural Networks (GNNs) \citep{kipf2017semisupervised,mpgnn,gat,graphsage} have emerged as the leading approach for such tasks. The fundamental idea behind GNNs is to use neural-networks that combine the node features with the graph-structure, in order to obtain useful graph representations. This combination is done in an iterative manner, which can capture complex properties of the graph and its node features. 

Although \structs{} are provided as input to the GNN, in some cases the best solution can be obtained by ignoring them. This may be due to these \structs{} being non-informative for the predictive task at hand. For instance, some molecular properties such as the molar mass (i.e., weight) depend solely on the constituent atoms (node features), and not on the molecular structure. 
Another case is when the provided \struct{} does contain valuable information for the task, but the GNN cannot effectively exploit it. In such cases, better test accuracy may be achieved by ignoring the \struct.
In other cases, the node features alone carry most of the information and the \struct{} conveys just a small added value. For example, assume that node features contain the zipcode of a user. Then the user's income is highly predictable by that feature, and their social structure will add little accuracy of this prediction.

\comment{\amirg{I don't think this last point (starting with "in other cases" is needed. Remove if space needed.}}

Motivated by this observation, we ask a core question in GNN learning: will GNNs work well in cases where it is better to ignore the \struct{} or will they overfit the \struct{}, resulting in reduced test accuracy?
Answering this question has several far-reaching practical implications. To illustrate, if GNNs lack the ability to discern when to disregard the graph, then providing a graph can actually hurt the performance of GNNs, and thus one must carefully re-think which graphs to provide a GNN.
On the other hand, if GNNs easily reject the structure when they fail to exploit it, then practitioners should attempt to provide a graph, even if their domain knowledge and expertise suggest that there is only a small chance it is informative.

We consider the common setting of over-parameterized GNNs. Namely, when the number of parameters the GNN uses is larger than the size of the training data. This is a very common case in deep-learning, where the learned model can fit any training data. Previous studies showed that despite over-parameterization, models learned using Gradient Descent (GD) often generalize well. Hence, it was suggested that the learning algorithm exhibits an implicit bias (e.g., low parameter norm) to avoid spurious models that happen to fit the training data \citep[e.g.,][]{zhang2017understanding,lyu2020gradient,NEURIPS2018_0e98aeeb,implcit_bias_separable_data}. 

Our focus is thus on the implicit bias of GNN learning, and specifically whether GNNs are biased towards using or not using the \struct. If the implicit bias is towards ``simple models'' that do not use the \struct{} when possible, then one would expect GNNs to be oblivious to the \struct{} when it is not informative. Our first empirical finding is that this is actually not the case. Namely, GNNs tend to {\em not} ignore the graph, and their performance is highly dependent on the provided \struct{}. Specifically, there are \structs{} that result in models with low test accuracy.

Next, we ask which properties of the learned graph distribution affect the GNN's ability to ignore the graph.
We empirically show that graphs that are regular result in more resilient GNNs. 
We then analyze the implicit bias of learning GNNs with gradient decent and prove that despite the ground truth function being ``simple" in the sense it does not use the graph, GNNs are not guaranteed to learn a solution that ignores the graph, even with infinite data.
We prove that as a result of their implicit bias, GNNs may fail to extrapolate. We then prove that within the family of regular graphs, GNNs are guaranteed to extrapolate when learning with gradient descent, and provide a sufficient condition for extrapolation when learning on regular graphs.

Finally, we empirically examine on real-world datasets if the properties of regular graphs are also beneficial in cases where the graph should not necessarily be ignored. We show that modifying the input graph to be ``more regular'' can indeed improve performance in practice.


We note that we focus on the implicit bias of GNNs, i.e., what GNNs \textit{actually} do when the graph \textit{should} be ignored. Understanding this bias can also shed light on the phenomenon of entanglement \cite{10.1145/3394486.3403076, seddik2022node, chen2020graph},  i.e., the intricate interplay between the graph structure and the node features.


\textbf{The main contributions of this work are} (1) We show that GNNs tend to overfit the \struct, when it should be ignored.
(2) We evaluate the \struct{} overfitting phenomenon with respect to different graph distributions and find that the best performance is obtained for regular graphs.
(3) We theoretically analyze the implicit bias of learning GNNs, and show that when trained on regular graphs, they converge to unique solutions that are more robust to \struct{} overfitting. 
(4) We show empirically that transforming GNN input graphs into more regular ones can mitigate the GNN tendency to overfit, and improve performance.



\section{GNNs Overfit the Graph-Structure}
In this section, we present an empirical evaluation showing that GNNs tend to overfit the \struct, thus hurting their generalization accuracy. Graph overfitting refers to any case where the GNN uses the graph when it is preferable to ignore it (e.g.,  because it is non-informative for the task).

\subsection{Preliminaries}
A graph example is a tuple $G = (A, X)$. $A$ is an adjacency matrix representing the \struct{}. Each node $i$ is assigned a feature vector $\x_i \in \reals^{d}$, and all the feature vectors are stacked to a feature matrix $X\in \reals^{n\times d}$, where $n$ is the number of nodes in $G$. The set of neighbors of node $i$ is denoted by $N(i)$. We denote the number of samples in a dataset by $m$.
We focus on the common class of Message-Passing Neural Networks~\cite{morris2021weisfeiler}.
In these networks, at each layer, each node updates its representation as follows:
\begin{equation}\label{eq:gnn}
    h_i^{(k)} = \sigma\left(W^{(k)}_1 h_i^{(k-1)} + W^{(k)}_2 \sum_{j\in N(i)} h_j^{(k-1)} +b^{(k)}\right) 
\end{equation}
where $W_1^{(k)}, W_2^{(k)} \in \reals^{d_k\times d_{k-1}}$. The initial representation of node $i$ is its feature vector $h_i^{(0)} = \x_i$.
The final node representations $\{h_i^{(L)}\}_{i=1}^{n}$ obtained in the last layer, can then be used for downstream tasks such as node or graph labeling.
We focus on graph labeling tasks, where a graph representation vector is obtained by combining all the node representations, e.g. by summation.
This is then followed by a linear transformation matrix $W_3$ that provides the final classification/regression output (referred to as a \textit{readout} layer).
For the sake of presentation, we drop the superscript in cases of one-layer GNNs.
For binary classification, we assume the label is the sign of the output of the network.
We refer to $W_1^{(k)}$ as the \textit{root-weights} of layer $k$ and to $W_2^{(k)}$ as the \textit{topological-weights} of layer $k$.
A natural way for GNNs to ignore the \struct{} is by zeroing the topological-weights $W_2^{(k)}=\bar{0}$ in every layer.
We say that a function $f(X,A)$ is \textit{graph-less}
if $f(X,A) = f(X)$, i.e., the function does not use the \struct, and is practically a set function.

It is important to note that some GNNs, e.g., \citet{gcn}, do not possess the ability to ignore the \struct{} as the root and topological weights are the same. We therefore focus on the most general GNN type that does have the ability to ignore the graph~\cite{mpgnn}.
In the Appendix, we extend our empirical evaluation to multiple GNN variations, including Graph Attention Network~\cite{gat, brody2022attentive}, Graph Transformer~\cite{shi2021masked} and Graph Isomorphism Network~\cite{gin}, and to node classification, which show similar trends. 


\begin{table}[t]
  \centering
      \caption{The accuracy of a fixed GNN architecture, trained once on the given graphs in the data (GNN) and once on the same data where the \struct{} is omitted ($GNN_{\emptyset}$), i.e., on empty graphs. The solution of $GNN_{\emptyset}$ is realizable by $GNN$, and the only difference between the runs is the given \structs{}. This suggests that the decreased performance of $GNN$ is due to \struct{} overfitting.}
     \label{table:topo_overfit}
\begin{tabular}{lccc}
\toprule
     & Sum&Proteins&Enzymes\\
\midrule 
$GNN$&94.5 $\pm$ 0.9& 67.4 $\pm$ 1.9& 55.2 $\pm$ 3.1 \\
$GNN_{\emptyset}$ &97.5 $\pm$ 0.7 &74.1  $\pm$ 2.5&  64.1 $\pm$ 5.7 \\
\bottomrule
\end{tabular}

\end{table}



 \begin{figure*}[t]
    \centering

      \subfigure[]{\label{figure:learning_curves_big} \includegraphics[width=0.45\textwidth]{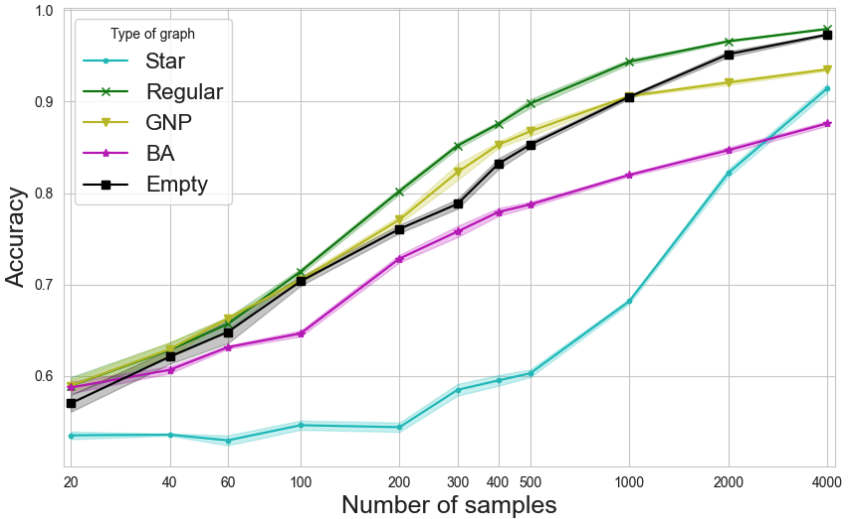} }\quad
      \subfigure[]{\label{figure:norm_ratio_big}\includegraphics[width=0.45\textwidth]{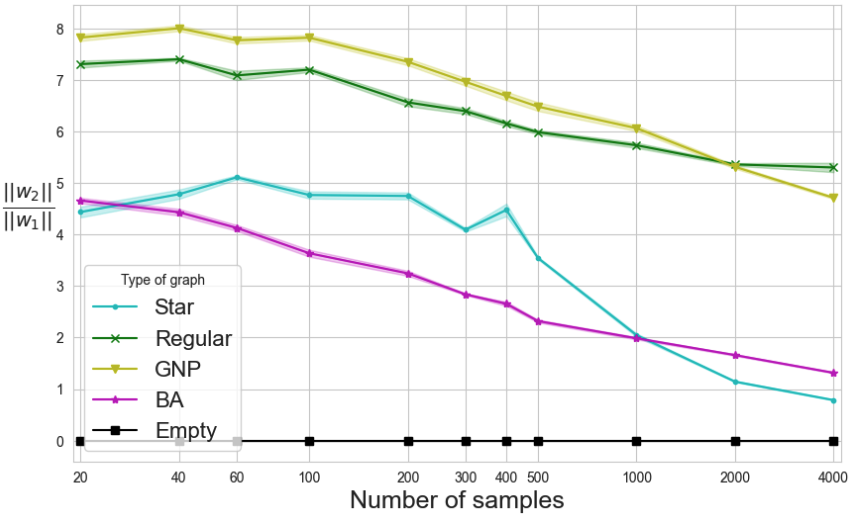}}

    \caption{(a) The learning curves of the same GNN model trained on graphs that have the same node features and only differ in their \struct{}, which is sampled from different distributions.  
    The label is computed from the node features without the use of any \struct{}.
    If GNNs were to ignore the non-informative \struct{} they were given, similar performance should have been observed for all graph distributions. Among the different distributions, regular graphs exhibit the best performance.
    (b) The norm ratio between the topological and the root weights along the same runs. Except for the empty graphs, the ratio is always greater than $1$, which indicates that more norm is given to the topological weights. }
    \label{Figure:sample_complexity_and_norms}
\end{figure*}

\subsection{Evidence for Graph Overfitting}{\label{section:graph_overfitting}}
Our goal is to examine what happens when GNNs learn over graphs that should be ignored, either because they are non-informative for the task, or because the GNN fails to exploit their information. 
To that end, we conducted experiments on three datasets.

\paragraph{Sum}
This is a binary classification synthetic task, with a graph-less ground truth function. To generate the label, we use a teacher GNN that simply sums the node features, and applies a linear readout to produce a scalar.
The data contains non-informative \structs{} which are drawn from the GNP graph distribution~\cite{gnp}, where the edges are sampled i.i.d with probability $p$ (we used $p=0.5$).
\newline
\newline
\textbf{Proteins and Enzymes}
These are two classification tasks on real-world molecular data~\cite{tudataset}.
In~\citet{errica2022fair} the authors reported on a thorough GNNs comparison, that the best accuracy on these datasets is achieved when the \struct{} is omitted.
We note that with a fixed architecture, the solution learned by a GNN trained on empty graphs is always realizable by the same GNN trained on non-empty graphs. This is a straight-forward argument and it is explained in the Appendix for the sake of completeness.
Therefore, with a fixed architecture, better performances that are achieved when learning over empty graphs indicates that it was better for the GNN to ignore the graph, and it could, but it didn't. 
\citet{errica2022fair} used a different model for the empty graphs, which was not an instance of the other compared GNNs. Therefore, their results does not imply that the compared GNNs overfitted the \struct, as the superiority of the model trained on empty graphs may be due to its architecture.
In our experiments, we use a fixed architecture to ensure that a discrepancy in perforamance implies graph overfitting.

\paragraph{Protocol and Results} 
On each of the three datasets, we trained the same GNN twice: once on the given \struct{}s in the data ($GNN$), and once when the \struct{} is replaced with an empty graph and only the node features are given for training ($GNN_{\emptyset}$). 
This difference between these setups shows the effect of providing the \struct.

The GNNs architecture is fixed and the learning hyper-parameters are tuned on a validation set for the Sum task, and $10$-fold cross-validation for Protein and Enzymes. We report test errors averaged over $10$ runs with random seeds on a separate holdout test set. More information can be found in the Appendix. 

Table~\ref{table:topo_overfit} shows the results of the experiments. In the three tasks, $GNN_{\emptyset}$ achieves higher accuracy than $GNN$.
This suggests that $GNN$ made use of the graphs, although a better result, i.e., the one learned by $GNN_{\emptyset}$, could be obtained by ignoring them. This {\em graph overfitting} led to lower test accuracy.

\subsection{How Graph-Structure Affects Overfitting}\label{section:different_dist_overfitting}
The previous section showed that in the Sum task, where the given \structs{} are non-informative and should be ignored, the GNN overfits them instead. 
Here we further study how this phenomenon is affected by the specific graph-structure provided to the GNN. Thus, we repeat the setup of the Sum task but with different graph distributions.

\paragraph{Data} 
We used the Sum task described in Section~\ref{section:graph_overfitting}.
We created four different datasets from this baseline, by sampling graph-structures from different graph distributions. The set of node feature vectors remains the same across all the datasets, and thus the datasets differ only in their \structs.
The graph distributions we used are: $r$-regular graphs (Regular) where all the nodes have the same degree $r$, star-graph (Star) where the only connections are between one specific node and all other nodes, the Erd\"os-R\'enyi graph distribution (GNP) ~\cite{gnp},  where the edges are sampled i.i.d with probability $p$, and the preferential attachment model (BA) ~\cite{badist}, where the graph is built by incrementally adding new nodes and connecting them to existing nodes with probability proportional to the degrees of the existing nodes.

\paragraph{Protocol} 
The GNN model is as in the Sum task in the previous section.
On each dataset, we varied the training set size and evaluated test errors on $10$ runs with random seeds.
More information can be found in the Appendix.

\paragraph{Results}
For the sake of presentation, we present the results on one instance from each distribution: Regular with $r=10$, GNP with $p=0.6$ and BA with $m=3$.
Additional results with more distribution parameters are given in the Appendix and show similar trends.
Recall that the datasets differ only by the edges and share the same set of nodes and features. 
Therefore, had the GNN ignored the \structs{}{}, we would expect to see similar performance for all datasets.
As shown in Figure~\ref{figure:learning_curves_big}, the performance largely differs between different graph distributions, which indicates the GNN overfits the graphs rather than ignores them. 

To further understand what the GNN learns in these cases, we evaluate the ratio between the norms of the topological and root weights. Results are shown in Figure~\ref{figure:norm_ratio_big}.
It can be seen that for all the graphs except the empty graphs, the ratio is larger than $1$, indicating that there is more norm on the topological weights than on the root weights. Specifically, the \struct{} is not ignored.  In the case of empty graphs, the topological weights are not trained, and the ratio is $0$ due to initialization. We also present the norms of the root and topological weights separately in the Appendix.

Figure~\ref{Figure:sample_complexity_and_norms} suggests that some graph distributions are more robust to \struct{} overfitting. The GNN trained on regular graphs performs best across all training set sizes. 

The good performance on regular graphs would seem to suggest that it learns to use low topological weights. However as Figure~\ref{figure:norm_ratio_big} shows, the opposite is  actually true.
This may seem counter-intuitive, but in the next section we theoretically show how this comes about.




\section{Theoretical Analysis}

In the previous section, we saw that GNNs tend to overfit the \struct{} when it should be ignored.
We now turn to a theoretical analysis that sheds light on what GNNs learn when the ground truth teacher is graph-less. 


For the sake of clarity, we state all theorems for a one-layer GNN with sum-pooling, no readout, and output dimension $1$. For simplicity, we also assume no bias term in our analysis. All the proofs and extensions can be found in the Appendix. 

\subsection{Implicit bias of Gradient Descent for GNNs}
Let $S$ denote a training set of labeled graphs. Each instance in $S$ is a triplet $(X,A,y)$, where $X$ is a stacked feature matrix of the node feature vectors, $A$ is the adjacency matrix, and $y\in\pm 1$ is the class label (we consider binary classification).
To examine the solutions learned by GNNs, 
we utilize Theorem 4 from \citet{NEURIPS2018_0e98aeeb}. This theorem states that homogeneous neural networks trained with GD on linearly separable data converge to a KKT point of a max-margin problem.
Translating this theorem to the GNN in our formulation, we get  that gradient-based training will converge to the solution of the following problem:
\begin{equation}{\label{eq:max_margin_general}}
    \begin{array}{l}
     \min_{\w_1,\w_2}   \|\w_1\|_2^2 + \|\w_2\|_2^2  \\
    s.t. ~~~ y [ \w_1 \cdot \sum_i^n \x_i + \w_2 \sum_{i}^n deg(i) \x_i]  \geq 1 \\
     \forall (X,A,y)\in S 
    \end{array}
\end{equation}

Equation~\ref{eq:max_margin_general} can be viewed as a max-margin problem in $2d$ space, where the input vector is $[\sum_i^n \x_i , \sum_i^ndeg(i)\x_i]$.
Therefore, the graph input can be viewed as the sum of the node feature vectors concatenated with their weighted sum, according to the node degrees. 

When trained on $r$-regular graphs, Equation~\ref{eq:max_margin_general} can be written as:
\begin{equation}{\label{eq:max_margin_regular}}
    \begin{array}{ll}
    \min_{\w_1,\w_2} & \|\w_1\|_2^2 + \|\w_2\|_2^2  \\
    s.t. & y [(\w_1 + r\w_2) \cdot \sum_i^n \x_i]  \geq 1 \\ \forall (X,A,y)\in S 
    \end{array}
\end{equation}
This can be viewed as a max-margin problem in $\mathbb{R}^d$ where the input vector is $ \sum_i^n\x_i$. So the GNN is a linear classifier on the sum of the node features, but the regularizer is not the $L_2$ norm of the weights, because of the $r$ factor.


The next theorem shows that when a GNN is trained using
GD on regular graphs, the learned root and topological weights are aligned.

\begin{lemma}[Weight alignment]\label{thm:weight_aligment}
Let $S$ be a set of linearly separable $r$-regular graph examples. A GNN trained with GD that fits $S$ perfectly converges to a solution such that $\w_2 = r\w_1$. Specifically, the root weights $\w_1$ and topological weights $\w_2$ are aligned.
\end{lemma}

We prove Lemma \ref{thm:weight_aligment} in the Appendix by analyzing the KKT conditions for first order stationary points of Equation~\ref{eq:max_margin_regular}.


The next section will use this result to explain why  regular graphs are better for learning graph-less teachers.
\subsection{Extrapolation with graph-less teachers}
In this section, we analyze what
happens when GNNs learn from training data generated by
a graph-less model (we refer to this as a graph-less teacher).
As we saw empirically in Section~\ref{section:graph_overfitting}, these learned models will sometimes
generalize badly on test data. We begin with a theorem
that proves that such bad cases indeed exist. The theorem
considers the extrapolation case, where the train and test distribution of
graphs is not the same, but is labeled by the same graph-less
teacher. Had the GNN learned a graph-less model, it would
have had the same train and test performance (for infinite
data). However, we show this is not the case, indicating that GNNs can overfit the graph structure arbitrarily badly. In other words, they do not extrapolate.



\begin{theorem}[Extrapolation may fail]{\label{thm:extrapolation_fail}}
 Let $f^*$ be a graph-less teacher. There exist graph distributions $P_1$ and $P_2$, with node features drawn from the same fixed distribution, such that when learning a linear GNN with GD over infinite data drawn from $P_1$ and labeled with $f^*$, the test error on $P_2$ lableled with $f^*$ will be $\geq \frac{1}{4}$. Namely, the model will fail to extrapolate.
\end{theorem}
The setting in the above result is that a graph-less ground truth teacher is learned using graphs from $P_1$. Ideally, we would have liked GD to ``ignore'' the graphs, so that the output of the learned model would not change when changing the support to $P_2$. However, our result shows that when the graph distribution is changed to $P_2$, performance is poor. This is in line with our empirical observations. The key idea in proving the result is to set $P_2$ such that it puts weights on isolated nodes, and thus exposes the fact that the learned function does not simply sum all nodes, as the graph-less solution does. 

Despite Theorem~\ref{thm:extrapolation_fail} showing GNNs may fail to extrapolate, the following result shows that GNNs are guaranteed to extrapolate within the family of regular distributions. 
\comment{\amirg{this also needs infinitely many samples. Otherwise of course you can fail to also interpolate. Need to change wording here. See proposal below}
\amirg{
Proposal for formulation:
Let $D_G$ be a distribution over r-regular graphs and $D_X$ be a distribution over node features. Assume a training set of infinite size sampled from $D_G$ and $D_X$ and labeled with a graph-less teacher. Denote the learned model by $f$. 
Assume that test examples are sampled from $D'_G$, a distribution over r'-regular graphs, and $D_X$. Then $f$ will have zero test error, 
}}
\begin{theorem}[Extrapolation within regular distributions]\label{thm:ood_regular}
Let $D_G$ be a distribution over r-regular graphs and $D_X$ be a distribution over node features. Assume a training set of infinite size sampled from $D_G$ and $D_X$ and labeled with a graph-less teacher. Denote the model learned with GD by $f$. 
Assume that test examples are sampled from $D'_G$, a distribution over r'-regular graphs, and $D_X$. Then $f$ will have zero test error.
\end{theorem}
\comment{Let $S$ be a set of linearly separable graph examples drawn from a distribution over $r$-regular graphs, with binary labels. Assume that the teacher is graph-less. Then a GNN that fits $S$ perfectly will extrapolate to any distribution over $r'$-regular graphs for all values of $r'$.}

To prove Theorem~\ref{thm:ood_regular}, we utilize Equation \ref{eq:max_margin_regular} and Lemma \ref{thm:weight_aligment}, and show that the direction of the weight vector used by the GNN does not change when the regularity degree is changed.
It was previously shown in \citet{size_generalization} that when there is a certain discrepancy between the train and test distributions, GNNs may fail to extrapolate. The argument extends to our case, and therefore learning without GD could fail to generalize.
%



\comment{
We next show that when learning without GD, Theorem~\ref{thm:ood_regular} does not hold. In other words, there are solutions that fit the training set perfectly and will fail to extrapolate to any regular graph with a regularity degree different from the one of the training set. In the proof, we construct such a solution.

\begin{lemma}{\label{lemma:extrapolation_bad_sol}}
  Let $S$ be a set of examples drawn from an $r$-regular graphs distribution and labeled with a graph-less teacher. Then there is a GNN that will fit $S$ perfectly and will produce the wrong label for $S$ when its graphs are changed to $r'$-regular, $r'\neq r$.

\end{lemma}
It was previously shown in \citet{size_generalization} that when there is a certain discrepancy between the train and test distributions, GNNs may fail to extrapolate. Lemma~\ref{lemma:extrapolation_bad_sol} extends the setting of the result from \citet{size_generalization}.

}
\comment{
    Let $S$ be a set of linearly separable graph examples drawn from an $r$-regular graphs distribution, with binary labels. Assume that the ground truth function is graph-less.\amirg{let's avoid the linearly separable stuff. Just say that $S$ is labeled by a graph-less teacher.} Then there is a GNN that will fit $S$ perfectly and will produce the wrong label for any $r'$-regular graphs with $r'\neq r$.\amirg{Not clear what distributions over features we are talking about here and what is the train and test distribution.} 
    
\amirg{Alternative:
    Let $S$ be a set of examples drawn from an $r$-regular graphs distribution and labeled with a graph-less teacher. Then there is a GNN that will fit $S$ perfectly and will produce the wrong label for $S$ when its graphs are changed to $r'$-regular.
}}



\begin{figure}[t]
\centering
        \includegraphics[width=1\linewidth]{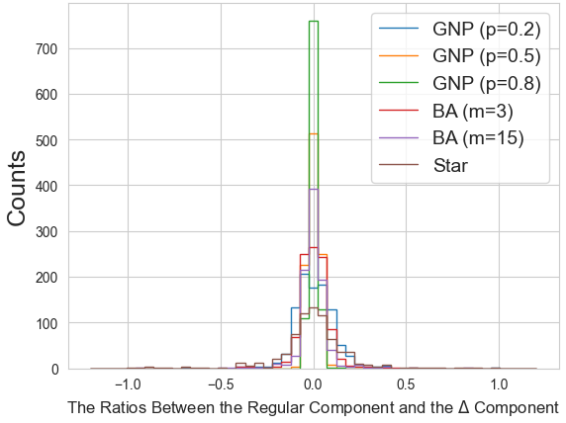}

        \caption{The ratios histogram for test examples that are correctly classified in the extrapolation evaluation presented in Table~\ref{table:regular_extrapolation}. The condition in Theorem~\ref{thm:extrapolation} is met for all the correctly classified examples.}
        \label{Figure:gnp_extrapolation}
\end{figure}

\subsubsection{Characterizing Extrapolation Accuracy}{\label{sec:extrapolation_sufficient_condition}}
 Theorems~\ref{thm:extrapolation_fail} and \ref{thm:ood_regular} show extreme cases of good and bad extrapolation. We next examine what determines the extrapolation accuracy.
 
First, we empirically observe that GNNs trained on regular graphs exhibit good extrapolation to other non-regular graph distributions as well,  as presented in Table~\ref{table:regular_extrapolation}. 
For example, a GNN trained on $5$-regular graphs, generalizes perfectly to GNP graphs, and there is a decrease in performance when tested on star-graphs.  The training setup and more information on the graphs can be found in the Appendix.

Next, we present a sufficient condition for extrapolation and empirically show on these test sets that indeed when the GNN successfully extrapolates, this sufficient condition holds.

We utilize Lemma~\ref{thm:weight_aligment} and write the GNN trained on $r$-regular graphs in a new form acting on a test graph $G$ as:
$$f(G) = 
\underbrace{W_1\sum_{i=1}^{n}\x_i + rW_1\sum_{i=1}^{n}r'\x_i}_{Regular~Component} + \underbrace{rW_1\sum_{i=1}^{n}\Delta_{r',G}(i)\x_i}_{\Delta~Component}
$$
where $\Delta_{r',G}(i) = deg_G(i)-r'$, for any $0 \leq r'$.
This notation shows that applying $f$ to a graph $G$ is equivalent to applying it to an $r'$-regular graph plus applying it to another $\Delta$-graph that depends on $r'$.

Using this notation, the following Theorem provides a sufficient condition for extrapolation. For simplicity, we state the results as extrapolation to the same training set, with modified graphs.

\comment{\amirg{There's an issue here. It seems like you are assuming that
this model correctly classifies all r-regular graphs, not just on the training set. e.g., assume that the training set is of size $1$. I added a revised version of this and changed the proof accordingly.}
\amirg{I suggest adding: For simplicity, we state the results as extrapolation to the same training set, with modified graphs.}}
\begin{table}[t]
  \centering
      \caption{Accuracy of a GNN trained on $5$-regular graphs and tested on different distribution shifts. The GNN extrapolates perfectly to regular graph distributions, as guaranteed by Theorem~\ref{thm:ood_regular}.}
      \label{table:regular_extrapolation}
    \begin{tabular}{lc}
\toprule
        Test distribution & Accuracy \\
\midrule
        Regular (r=10) & 100  $\pm$ 0.0   \\ 
        Regular (r=15) & 100 $\pm$ 0.0   \\
        GNP (p=0.2) & 100 $\pm$ 0.0  \\ 
        GNP (p=0.5) & 100  $\pm$ 0.0  \\ 
        GNP (p=0.8) & 100 $\pm$ 0.0  \\ 
        BA (m=3) & 98.0 $\pm$ 1.7  \\ 
        BA (m=15) & 93.2 $\pm$ 0.9  \\
        Star Graph & 75.9 $\pm$ 1.1  \\
        \bottomrule
    \end{tabular}

\end{table}


\begin{theorem}[Sufficient condition for extrapolation]{\label{thm:extrapolation}}

Let $S$ be a set of $r$-regular graphs examples, labeled with a graph-less teacher $f^*$. Let $f$ denote a GNN trained with GD on $S$.
Now assume an instance $G = (X,A)\in S$ has been modified to a different graph $\tilde{G}=(X,\tilde{A})$ such that  there exists an $0 \leq r' \leq n-1$ where: 
$$
\left|\frac{r\w_1\sum_{i=1}^{n}\Delta_{r',\tilde{G}}(i)\x_i}{\w_1\tilde{x} + r'r\w_1\tilde{x}}\right| \leq 1 
$$
Then $f(\tilde{G})=f^*(\tilde{G})$ .

\end{theorem}
\comment{ 
Let $S$ be a set of $r$-regular graphs examples, labeled with a graph-less teacher $f^*$. Then a GNN that fits $S$ perfectly will correctly classify a test graph $G$ labeled by $f^*$ if there exists an $0 \leq r' \leq n-1$ such that
$$
\left|\frac{r\w_1\sum_{i=1}^{n}\Delta_{r',G}(i)\x_i}{\w_1\tilde{x} + r'r\w_1\tilde{x}}\right| \leq 1 
$$
\amirg{
Alternative:
Let $S$ be a set of $r$-regular graphs examples, labeled with a graph-less teacher $f^*$. Let $f$ denote a GNN trained with GD on $S$.
Now assume an instance in $G = (X,A)\in S$ has been modified to a different graph $\tilde{G}=(X,\tilde{A})$ such that  there exists an $0 \leq r' \leq n-1$ where: 
$$
\left|\frac{r\w_1\sum_{i=1}^{n}\Delta_{r',\tilde{G}}(i)\x_i}{\w_1\tilde{x} + r'r\w_1\tilde{x}}\right| \leq 1 
$$
Then $f(\tilde{G})=f^*(\tilde{G})$ .
}
}

\comment{
\begin{theorem}[Sufficient Condition for Extrapolation]{\label{thm:extrapolation}}
Let $f$ be a GNN that perfectly fits a training set of $r$-regular graphs.
Given a test graph $G$ from some graph distribution, if there exists an $0 \leq r' \leq n-1$ such that
$$
\mid\frac{r\w_1\sum_{i=1}^{n}\Delta_{r',G}(i)\x_i}{\w_1\tilde{x} + r'r\w_1\tilde{x}}\mid \leq 1 
$$
Where $\Delta_{r',G}(i) = deg_G(i)-r'$,  then $f$ will classify $G$ correctly.
\end{theorem}
}
Theorem~\ref{thm:extrapolation} suggests that applying the GNN to graphs that are ``closer'' to regular graphs, i.e., have smaller $\Delta$, results in better extrapolation.
To prove it, we show that when these conditions hold,  the extrapolation is guaranteed from Theorem~\ref{thm:ood_regular}.

Next, we empirically show that indeed all the samples that were classified correctly in Table~\ref{table:regular_extrapolation} satisfy this condition of Theorem~\ref{thm:extrapolation}.

Figure~\ref{Figure:gnp_extrapolation} presents histograms of the values of the ratio in Theorem~\ref{thm:extrapolation} for every example that is correctly classified, over the test examples presented in Table~\ref{table:regular_extrapolation}.
We do not include regular graphs in the histograms, because extrapolation withing regular graphs is guaranteed from Theorem~\ref{thm:extrapolation}.
The ratio is computed for the $r'$ that minimizes the denominator of the ratio. 
Indeed, all the ratios are less than 1, and therefore the sufficient condition holds. These results demonstrate that indeed ``closeness to a regular'' graph is an important determinant in extrapolation accuracy. 



\section{Are Regular Graphs Better
when Graphs are Useful?
}{\label{sec:experiments}}

In the previous sections, we showed that regular graphs exhibit robustness to the tendency of GNNs to overfit non-informative graphs that should be completely ignored. 
In this section, we examine if regular graphs are also beneficial in scenarios when the graph may be informative. We perform an empirical evaluation on real-world data, where we do not know in advance if the graph is indeed informative or not. We compare the performance of the same method, when trained on the original graph, and on the same graph when transformed to be ``more regular".\footnote{The code is available on \url{https://github.com/mayabechlerspeicher/Graph_Neural_Networks_Overfit_Graphs}}


\begin{table*}[t]
 \centering
   \centering\fontsize{9}{11}\selectfont

    \caption{Performance of different GNNs when trained on the original graphs versus when the COV of the graphs is reduced. The best model is in bold and with an underline in cases where the p-value < 0.05 using the Wilcoxon signed-rank test.}
     \label{add_edge_real_data_exp}
 \bigskip


\begin{tabular}{llcccccc}
\toprule
     Model & Graph &  Proteins &NCI1 &Enzymes& D\&D & mol-hiv & mol-pcba  
\\
\midrule
DeepSet& Empty Graph & 74.1 ± 2.5 & 72.8 ± 2.1 &  64.2 ± 3.0 & 77.5 ± 2.0 & 69.5 ± 2.9 & 15.0 ± 0.6\\
\midrule
\multirow{2}{*}{GraphConv} 
& Original Graph & 73.1 ± 1.6 &76.5 ± 1.2 & 58.2 ± 2.1& 72.5 ± 1.7 &  78.2 ± 3.0 & 20.5 ± 0.5  \\
& Original Graph + R-COV  &\underline{ \textbf{75.5 ± 1.8}}  &\underline{ \textbf{80.1 ± 0.9}} & \underline{ \textbf{61.0 ± 1.5}} &  \textbf{74.8 ± 2.9} &  \underline{\textbf{80.9 ± 1.8}} & \underline{ \textbf{22.8 ± 0.5}}\\
\midrule
\multirow{2}{*}{GIN} 
& Original Graph & 72.2 ± 2.9& 79.2 ± 1.5 & 58.9 ± 1.8 & 74.5 ± 2.3 &  77.0 ± 1.9 & 21.1 ± 0.5 \\
& Original Graph + R-COV  & \underline{\textbf{74.8 ± 2.1}}  &\textbf{80.0 ± 1.1} & \textbf{59.7 ± 1.4} & \textbf{75.7 ± 3.9} &  \textbf{77.9 ± 1.3} & \textbf{21.5 ± 0.2}\\
\midrule
\multirow{2}{*}{GATv2} 
& Original Graph & 73.5 ± 2.8 & 80.4 ± 1.6& 59.9 ± 2.8& 70.6 ± 4.0 &  78.7 ± 2.5 & 23.5 ± 0.9\\
& Original Graph + R-COV  & \textbf{76.5 ± 2.0}& \underline{\textbf{83.0 ± 1.5}} & \underline{\textbf{63.9 ± 3.5}} &\textbf{73.9 ± 1.2} & \underline{\textbf{80.9 ± 2.0}} & \underline{\textbf{24.3 ± 0.7}}\\
\midrule
\multirow{2}{*}{GraphTransformer} 
& Original Graph & 73.9 ± 1.5& 80.5 ± 1.1 & 60.9 ± 2.1&74.1 ± 1.9 & 80.5 ± 2.9 & 29.1 ± 0.7\\
& Original Graph + R-COV  &  \underline{\textbf{76.7 ± 1.4}}& \underline{\textbf{83.1 ± 1.9}} &  \underline{\textbf{64.0 ± 1.9}} & \underline{\textbf{77.1 ± 1.8}} & \textbf{82.4 ± 1.5} & \underline{\textbf{30.5 ± 0.2}}\\
\bottomrule

\end{tabular}

\bigskip
\begin{tabular}{llcccccc}
\toprule
     Model & Graph &  IMDB-B& IMDB-M  & Collab & Reddit-B & Reddit-5k &  
\\
\midrule
DeepSet& Empty Graph& 70.0 ± 3.0 & 48.2 ± 2.5& 71.2 ± 1.3 & 80.9 ± 2.0 & 52.1 ± 1.7\\
\midrule
\multirow{2}{*}{GraphConv} 
& Original Graph & 69.6 ± 1.7& 47.5 ± 1.0 & 73.5 ± 1.3  &83.2 ± 1.5 & 50.0 ± 2.1 \\
& Original Graph + R-COV  & \underline{\textbf{72.9 ± 0.5}}& \underline{\textbf{50.0 ± 1.5}} & \textbf{74.2 ± 2.1} &  \underline{\textbf{87.0 ± 1.8}}  & \underline{\textbf{52.5 ± 1.7}} \\
\midrule
\multirow{2}{*}{GIN} 
& Original Graph & 70.1 ± 2.9  &  48.1 ± 2.5 & 75.3 ± 2.9 &89.1 ± 2.7 & 56.1 ± 1.5 \\
& Original Graph + R-COV  & \underline{\textbf{71.3 ± 1.5}}& \textbf{48.5 ± 1.7} & \underline{\textbf{77.2 ± 2.0}}  &\textbf{91.0 ± 1.1} & \textbf{56.7 ± 0.8} \\
\midrule
\multirow{2}{*}{GATv2} 
& Original Graph & 72.8 ± 0.9 & 48.4 ± 2.1 & 73.9 ± 1.7 & 90.0 ± 1.5& 56.4 ± 1.5 \\
& Original Graph + R-COV  & \underline{\textbf{75.8 ± 1.5}}& \underline{\textbf{50.8 ± 1.7}} & \textbf{75.1 ± 1.9} & \underline{\textbf{92.1 ± 0.9 }} & \textbf{57.0 ± 0.9} \\
\midrule
\multirow{2}{*}{GraphTransformer} 
& Original Graph & 73.1 ± 1.3& 49.0 ± 1.9 & 73.8 ± 1.5 & 90.6 ± 1.3 & 51.4 ± 1.7 \\
& Original Graph + R-COV  &  \underline{\textbf{76.1 ± 2.0}}& \textbf{51.1 ± 2.3} & \underline{\textbf{76.0 ± 1.8}}  &\underline{\textbf{92.3 ± 1.0}} & \underline{\textbf{ 56.0 ± 1.2}} \\
\bottomrule

\end{tabular}

\end{table*}

\paragraph{Setup}
Ideally, we would like to examine the performance change when a graph is transformed into a regular graph.
While it is possible to make a graph regular by simply completing it into a full graph, this approach may be computationally expensive or even infeasible when learning with GNNs.
Unfortunately, other approaches may require removing edges from the original graph, which may lead to information loss or even make the task non realizable.
Therefore, we evaluate a transformation of a given graph to a ``more regular" graph, while keeping all its original information. 
We modify a given graph by adding edges between low degree nodes in order to reduce its node degrees coefficient of variation (COV), i.e., the ratio between the standard-deviation and mean of the node degrees. The COV of regular graphs is $0$, because all the nodes have the same degree.
In order to maintain the original information about the given graph we augment each edge with a new feature that has values  $1$ and $0.5$ for the original and added edges, respectively. 

We evaluate\footnote{The code is provided in the Supplementary Material.} the four GNN architectures for which Section~\ref{section:graph_overfitting} shows a tendency to overfit the graph. These are: GraphConv~\cite{mpgnn}, GIN~\cite{gin}, GATv2~\cite{gat, brody2022attentive} and GraphTransformer~\cite{shi2021masked}.
We adopted the neighbors' aggregation component of each model to process the edge features in a non-linear way.
The complete forms of each architecture we used with the addition of edge feature processing can be found in the Appendix.

\subsection{Datasets} 
We used $11$ graph datasets, including two large-scale datasets, which  greatly differ in their average graph size and density, the number of node features, and the number of classes.

\textbf{Enzymes, D\&D, Proteins, NCI1}~\cite{wl} are datasets of chemical compounds where the goal is to classify each compound into one of several classes. \\
\textbf{IMDB-B, IMDB-M, Collab, Reddit-B, Reddit-5k}~\cite{dgk} are social network datasets.\\
\textbf{mol-hiv, mol-pcba}~\cite{ogb} are large-scale datasets of molecular property prediction.

More information on the datasets and their statistics can be found in the Appendix.

\paragraph{Evaluation}
For each model and for each task, we evaluate the model twice: on the original graph provided in the dataset (Original Graph) and on the original graph with the COV reduced (R-COV).
Because different graphs have different COVs, we set COV to a fixed percentage of the original average COV of each dataset separately. The percentage is a hyprparameter, and we tested the values $\{80\%, 50\%\}$.
We also include as a baseline the performance when the graph-structure is omitted (Empty Graphs) which is equivalent to using DeepSets~\cite{zaheer2018deep}.

For all the datasets except mol-hiv and mol-pcba we used $10$-fold nested cross validation with the splits and protocol of~\citet{errica2022fair}.
The final reported result on these datasets is an average of $30$ runs ($10$-folds and $3$ random seeds).
The mol-hiv and mol-pcba datasets have pre-defined train-validation-test splits and metrics \citet{ogb}. The metric of mol-hiv is the test AUC averaged over $10$ runs with random seeds. The metric of mol-pcba the metric is the averaged precision (AP) over its $128$ tasks.\\
Additional details and the hyper-parameters are provided in the Appendix.

\paragraph{Results}
Across all datasets and all models, reducing the COV of the graphs improves generalization.
Particularly intriguing outcomes are obtained in the PROTEINS and IMDB-M datasets. Within these two datasets, superior performance is attained when learning over empty graphs in comparison to the provided graphs. Nonetheless, reducing the COV improves performance also with respect to the empty graphs. This observation suggests that the structural information inherent in the data is indeed informative, yet the GNN fails to exploit it correctly as it is.

\begin{figure}[t]
    \centering
    \includegraphics[width=1\linewidth]{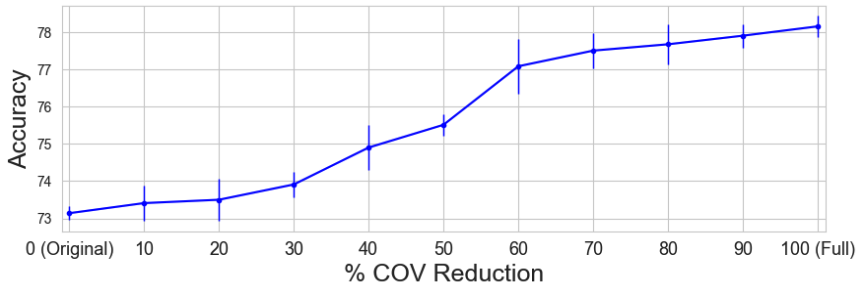}
    \caption{Accuracy and error bars of the Proteins datasets as the COV reduces. The performance is monotonically improving.}
    \label{Figure:proteins_rcov}
\end{figure}

\paragraph{The tradeoff between COV and graph density}
As we see consistent improvement when the COV is reduced, we further examined if this improvement is monotone with respect to the COV reduction. We evaluated the Proteins dataset with an increasing percentage of COV reduction, up to the full graph.
Indeed as shown in Figure~\ref{Figure:proteins_rcov}, the performance keeps improving as the COV is reduced. This is in alignment with the results of \citet{alon2021bottleneck} where a full-graph was used in the last layer of the network to allow better information flow between nodes of long distance. Note that in our case we also distinguish the original edges with the added edges using edge features, and allow the network to ignore the added edges. Clearly, using a full graph comes with a computational cost, a problem that also arises when using full-graph transformers.
Our results suggested that improvement in generalization can be achieved also without the cost of using the full graph. Practically,  one can limit the percentage of reduced COV according to their computation limit in advance.

\section{Practical Implications}

In practice, possible graph structures are typically determined based on domain knowledge, and it is common to explore multiple possible structures.
In some cases, a natural \struct{} inherently exists within the data, such as in social networks, where the network connections naturally define the graph layout. Nevertheless, it is usually not clear in advance if these graph layouts are informative for the task, or if the GNN will manage to exploit them.
The fact that certain layouts may provide valuable information for the task while others might not, and this distinction isn't clear beforehand, was the driving question for our research.
Indeed we found that the definition of the \struct, typically determined by users, emerges as a pivotal factor in performance outcomes due to the tendency of GNNs to overfit the provided graph.
This revelation opens up a fascinating avenue for further research into the significance of topological information during the training of GNNs. Understanding how GNNs respond to different structural layouts and why certain \structs{} are more effective than others could significantly impact the way we design and train these models.

\section{Future Work}
We believe this work opens up many new avenues for exploration.
One simple takeaway from our paper is to always try learning a model over empty graphs as well, i.e., using DeepSets~\cite{zaheer2018deep}. 
When the graph is known to have little contribution to the task, regularizing the topological weights may be useful.
The main difficulty is finding ways to improve the GNN's ability to exploit useful information from the graph if it exists and ignore it otherwise, without prior knowledge.
While we show in Section~\ref{sec:experiments} that reducing the graph's COV can enhance performance, there may be other ways to mitigate the graph overfitting. 
In recent years many methods were introduced to mitigate different phenomena that limit GNNs performance ~\cite{ Rong2019DropEdgeTD, alon2021bottleneck}. It is interesting to examine whether these methods are useful in mitigating graph overfitting.
Another interesting avenue for future research is analyzing the implicit bias of non-linear GNNs, including Graph Attention and Transformers.
\section{Conclusion}

In this study, we showed that although GNNs have the ability to disregard the provided graph when needed, they don't.  Instead, GNNs tend to overfit the \structs, which results in reduced performance. 
We theoretically analyzed the implicit bias of gradient-descent learning of GNNs, and proved that even with infinite data, GNNs are not guaranteed to learn a solution that ignores the graph, when the graph should be ignored.
We showed that regular graphs are more robust to graph overfitting, and provided a theoretical explanation and extrapolation results for this setting.
Our study shows that in some cases, the graph structure hurts the performance of GNNs, and therefore graph selection is of great importance, as well as having a model that can ignore the graph when needed.

\section{Acknowledgements}
This work was supported by a grant from the Tel Aviv University Center for AI and Data Science (TAD) and by the Israeli Science Foundation research grant 1186/18.

\bibliography{references}
\bibliographystyle{icml2024}
\clearpage
\appendix

\onecolumn

\section{Proofs and Extensions}
All our analysis assumes that train and test data are labeled via some graph-less teacher. Namely, a function $f^*$ that classifies a graph instance based only on its features and not the graph. We let $f^*$ be defined via a weight vector $\w_1^*$ as follows:
\begin{equation}
f^*(X) = \w^*_1\cdot \sum_{i=1}^{n}\x_i
\label{eq:graphless_func}
\end{equation}

For the sake of simplicity, we assume that all the hidden states are of dimension $d$.
We denote the number of vertices with $n$, the number of samples with $m$, and denote $\tilde{\x}= \sum_i^n \x_i$ for a set of node feature vectors $x_i\in G, 1\leq i \leq n, 1\leq l \leq m$.
$\x_i^{(l)}$ is the feature vector of node $i$ in the graph sample $l$.
\comment{\amirg{you may also want to define the graphless teacher here $f^*$ since you refer to it in many places. You can say that we assume this teacher throughout the proofs.
Can be like:
All our analysis assumes that train and test data are labeled via some graph-less teacher. Namely a function $f^*$ that classifies a graph instances based only on its features adn not the graph. We let $f^*$ be defined via a weight vector $\w_1^*$ as follows:
\begin{equation}
f^*(X) = \w^*_1\cdot \sum_{i=1}^{n}\x_i
\label{eq:graphless_func}
\end{equation}
}}

\subsection{Proof of Lemma 3.1}
For the sake of simplicity, we begin by proving the simplest case of a GNN with one layer and no readout. Then we extend the proof to the case of readout and multiple layers. 
In $r$-regular graphs, $deg(i)=r$ for all nodes $i\in n$. Therefore Equation~\ref{eq:max_margin_general} can be written as:
\begin{equation*}
    \begin{array}{lll}
    \min_{\w_1,\w_2} & \|\w_1\|_2^2 + \|\w_2\|_2^2  \\
    s.t. & y^{(l)} [(\w_1 + r\w_2) \cdot \tilde{\x}^{(l)}]  \geq 1 
    & \forall (X^{(l)},A^{(l)},y^{(l)})\in S 
    \end{array}
\end{equation*}

Now writing the KKT stationarity condition:

\begin{equation*}
\begin{aligned}
&\mathcal{L}(\w, \alpha) = \frac{1}{2}(\|\w_1\| + \|\w_2\|) - 
  \sum_{l=1}^m \alpha_l[y^{(l)}(\w_1+r\w_2)\tilde{\x}^{(l)}- 1] \\
  & \nabla _{\w} \mathcal{L} = 0 \iff \\
 &\frac{\partial \mathcal{L}}{\partial \w_1} = \w_1 -  \sum_{l=1}^m \alpha_ly^{(l)}\tilde{\x}^{(l)}= 0  
 \implies \w_1 =   \sum_{l=1}^m \alpha_ly^{(l)}\tilde{\x}\\
 &\frac{\partial \mathcal{L}}{\partial \w_2} = \w_2 -  r\sum_{l=1}^m \alpha_ly^{(l)}\tilde{\x}^{(l)} = 0
 \implies \w_2 =  r\sum_{l=1}^m \alpha_ly^{(l)}\tilde{\x}^{(l)}
\end{aligned}
\end{equation*}

Therefore  $\w_2 = r\w_1$. \qedsymbol

\paragraph{With Readout}
Assume a readout $W_3$ is applied after the sum-pooling, and denote $W=[W_1, W_2, W_3]$.
\begin{equation*}
\begin{aligned}
    &f(X, A, W) = W_3\sum_i^n h_i^{(2)} =
    W_3W_1\cdot \sum_i^n \x_i + W_3W_2\cdot \sum_i^{n} \sum_{j\in N(i)} \x_j =  \left(W_3W_1 \cdot + rW_3W_2 \right)\tilde{\x}
\end{aligned}
\end{equation*}
Then similarly to Equation~\ref{eq:max_margin_regular}, the max-margin problem becomes:

\begin{equation*}
    \begin{array}{lll}
    \min_{W_1,W_2, W_3} & \|W_1\|_{F}^2 + \|W_2\|_{F}^2 + \|W_3\|_F^2  \\
    s.t. & y^{(l)}\left[( W_3W_1 \cdot + rW_3W_2 )\tilde{\x}^{(l)}\right]  \geq 1 &   \forall (X^{(l)},A^{(l)},y^{(l)})\in S 
    
    \end{array}
\end{equation*}

Then the KKT stationarity condition:: 
\begin{equation*}
    \begin{aligned}
        & \mathcal{L}(W, \alpha) = \frac{1}{2}(\|W_3\|_F^2 +  \|W_1\|_{F}^2 + \|W_2\|_{F}^2 )-
        \sum_{l=1}^m \alpha_l \left( y^{(l)} W_3 \left[ \sum_{i=1}^{n} \left( W_1 \x_i^{(l)} + 
        r W_2 \x_i^{(l)} \right) \right] -1 \right) \\
        & \frac{\partial \mathcal{L}}{\partial W_1} = W_1 - \sum_{l=1}^m \alpha_l y^{(l)} \sum_{i=1}^{n} W_3 \x_i^{(l)} \\
        & \frac{\partial \mathcal{L}}{\partial W_2} = W_2 - r\sum_{l=1}^m \alpha_l y^{(l)} \sum_{i=1}^{n} W_3 \x_i^{(l)} \\
        & \frac{\partial \mathcal{L}}{\partial W_3} = W_3 - \sum_{l=1}^m \alpha_l y^{(l)} \sum_{i=1}^{n} \left( W_1 \x_i^{(l)} + r W_2 \x_i^{(l)} \right)\\
        & \nabla _{W} \mathcal{L} = 0 \iff \\
        & W_1 = W_3\sum_{l=1}^m \alpha_l y^{(l)}\sum_{i=1}^{n} \x_i^{(l)} \\ 
        & W_2 = rW_3\sum_{l=1}^m \alpha_l y^{(l)} \sum_{i=1}^{n}    \x_i^{(l)}\\
        & W_3 = \sum_{l=1}^m \alpha_l y^{(l)} \left( W_1 +rW_2\right) \left(\sum_{i=1}^{n}\x_i^{(l)}\right)
    \end{aligned}
\end{equation*}

Therefore $W_1$ and $W_2$ are aligned, as well as  $W_3W_1$ and $W_3W_2$ \comment{\amirg{I guess you don't mean the product $W_3W_1$ but rather $W_3,W_1$}\maya{I do mean the product}.}
\qedsymbol

\paragraph{Two Layers}
The updates in a 2-layer GNN are:

\begin{equation*}
    \begin{aligned}
    & \h_i^{(1)} = W_1^{(0)}\cdot \x_i + W_2^{(0)}\sum_{j\in N(i)} \x_j \\
    & \h_i^{(2)} = W_1^{(1)}\cdot \h^{(1)}_i + W_2^{(1)}\sum_{j\in N(i)} \h^{(1)}_j
    \end{aligned}
\end{equation*}
Then the final predictor is:
\begin{equation*}
    \begin{aligned}
    & f(X, A, W) = W_3\left(W_1^{(1)}W_1^{(0)} + \left(W_1^{(1)}W_2^{(0)} + W_2^{(1)} W_1^{(0)}\right)r 
      + W_2^{(1)}W_2^{(0)}r^2\right)\sum_i^n \x_i
    \end{aligned}
\end{equation*}

Therefore, let $W=[W_1^{(0)},W_2^{(0)},W_1^{(1)},W_2^{(1)},W_3]$ and we define: 
$$P(W) = W_3\left(W_1^{(1)}W_1^{(0)} + \left(W_1^{(1)}W_2^{(0)} + W_2^{(1)} W_1^{(0)}\right)r +
 W_2^{(1)}W_2^{(0)}r^2\right)$$
In this case Equation~\ref{eq:max_margin_general} becomes the following max-margin problem:
\begin{equation}\label{eq:2-layerM-max-margin}
    \begin{array}{lll}
    \min_W & \|W_1^{(0)}\|_F^2 + \|W_2^{(0)}\|_F^2 + \|W_1^{(1)}\|_F^2 + \|W_2^{(1)}\|_F^2 + \|W_3\|_F^2  \\
    s.t. & y^{(l)}\left[ P(W) \cdot \tilde{\x}^{(l)} \right]  \geq 1 & \forall (X^{(l)},A^{(l)},y^{(l)})\in S 
    \end{array}
\end{equation}

Using the KKT stationarity condition:

\begin{equation*}
    \begin{aligned}
    & \mathcal{L}(W, \alpha) = \frac{1}{2} \left(\sum
    \limits_{\substack{t\in \{0,1\}\\ k\in \{0,1\}}}
     \|W_t^{(k)}\|_F^2 + \|W_3\|_F^2 \right)
    - \sum_{l=1}^m\alpha_l[yP(W) \cdot \tilde{\x}^{(l)} - 1] \\
    & \frac{\partial \mathcal{L}}{\partial W_j} = W_j - \sum_{l=1}^m\alpha_ly \frac{\partial}{\partial W_j} (P(W) \cdot \tilde{\x}^{(l)}_i) \\
    & \frac{\partial}{\partial W_j} (P(W) \cdot \tilde{\x}^{(l)}) = \tilde{\x}^{(l)} \frac{\partial P(W)}{\partial W_j}  \\
    &  \frac{\partial \mathcal{L}}{\partial W_j} =0 \implies W_j = \sum_{l=1}^m\alpha_ly^{(l)} \tilde{\x}^{(l)}_i \frac{\partial P(W)}{\partial W_j}\\
    \\
    & \frac{\partial P(W)}{\partial W_1^{(0)}} = \frac{\partial}{\partial W_1^{(0)}} [W_3(W_1^{(1)} + rW_2^{(1)})W_1^{(0)}] \\
    & \frac{\partial P(W)}{\partial W_2^{(0)}} =  \frac{\partial}{\partial W_2^{(0)}} [W_3(rW_1^{(1)} + r^2W_2^{(1)})W_2^{(0)}] 
    =r\frac{\partial P(W)}{\partial W_1^{(0)}}\\
    \\
    & \frac{\partial  P(W)}{\partial W_1^{(1)}} = \frac{\partial}{\partial W_1^{(1)}} [W_3W_1^{(1)}(W_1^{(0)} + W_2^{(0)}r)] \\
    & \frac{\partial  P(W)}{\partial W_2^{(1)}} = \frac{\partial}{\partial W_2^{(1)}} [W_3W_2^{(1)}(W_1^{(0)} r + W_2^{(0)}r^2)] 
    = r\frac{\partial P(W)}{\partial W_1^{(1)}}\\
    \\
    \end{aligned}
\end{equation*}

Therefore the root and topological weights are aligned, in every layer.  \qedsymbol

\paragraph{L Layers - Overview }  In the case of L layers, the same holds, only $P(W)$ is different. We provide the sketch of the proof.
The max-margin problem is
\begin{equation}\label{eq:multiple_layers}
    \begin{array}{lll}
    \min_W & \|W_3 \|_F^2 + \sum_{k=0}^{L-1} \|W^{(k)}_0 \|_F^2  + \| W^{(k)}_1 \|_F^2 \\
    \\
    s.t. & y^{(l)}[P(W) \tilde{\x}^{(l)}]  \geq 1 & \forall (X^{(l)},A^{(l)},y^{(l)})\in S \\
    \end{array}
\end{equation}

Then the KKT stationarity condition:

\begin{equation*}
    \begin{aligned}
        & \mathcal{L}(W,\alpha) = \|W_3 \|_F^2 + \sum_{k=0}^{L-1} \|W^{(k)}_0 \|_F^2  + \| W^{(k)}_1 \|_F^2 -
        \sum_{l=1}^m \alpha_l [ y^{(l)} P(W) \tilde{\x}^{(l)} - 1 ] \\
        & \frac{\partial \mathcal{L}}{\partial W^{(k)}_j}  = 2W^{(k)}_j - \sum_{l=1}^m \alpha_l y^{(l)} \frac{\partial}{\partial W^{(k)}_j} \left(P(W)\tilde{\x}^{(l)}\right)\\
        & \frac{\partial \mathcal{L}}{\partial W^{(k)}_j} = 0 \implies W^{(k)}_j = \frac{1}{2} \frac{\partial P(W) }{\partial W^{(k)}_j} \cdot \sum_{l=1}^m \alpha_l y^{(l)} \tilde{\x}^{(l)}  
    \end{aligned}
\end{equation*}

Therefore, to show alignment between the root and topological weights in layer $k$, it is enough to show that 
\begin{equation*} \label{diff-alignment}
   r \frac{\partial P(W)}{ \partial W^{(k)}_0 } = \frac{\partial P(W)}{ \partial W^{(k)}_1 } 
\end{equation*}

\subsection{Proof of Theorem 3.2}
Here we prove Theorem 3.2 by showing providing distributions P1 and P2 such that GNNs trained on graphs from P1, will fail to extrapolate to graphs from P2. We consider the case where P1 is a distribution over r-regular graphs and P2 is a distribution over star graphs with a random center node. The key intuition in our proof is that learning with P1 will learn a model that averages over nodes. But when testing it on P2,  mostly the center node will have to determine the label, and this will typically result in an error. We will take the graph size to $\infty$ to simplify the analysis, but results for finite graphs with high probability can be obtained using standard concentration results.
\comment{
\amirg{This isn't clearly mapped to the theorem. I don't see $P1,P2$ I don't see a test error and I don't see $0.25$...
e.g. you can write Here we prove Theorem 3.2 by showing providing distributions P1 and P2 such that GNNs trained on graphs from P1, will fail to extrapolate to graphs from P2. We consider the case where P1 is a distribution over r-regular graphs and P2 is a distribution over star graphs with a random center node. The key intuition in our proof is that learning with P1 will learn a model that averages over nodes. But when testing it on P2,  mostly the center node will have determine the label, and this will typically result in an error. We will take the graph size to $\infty$ to simplify the analysis, but results for finite graph with high probability can be obtained using standard concentration results.}}
Let $f^*$ be a graph-less function,  $f^*(X) = w^*_1\sum_{i=1}^{n}x_i$.
We assume that $f^*$ labels the training graphs, and graphs are drawn from $P1$, namely a distribution over r-regular graphs.
Let $f$ be the learned function when trained with infinite data on $r$-regular graphs, with node features with dimension $1$ drawn from $\mathcal{N}(0,1)$.
Then $f(G) = sign(w_1\sum_{i=1}^{n}x_i+rw_1\sum_{i=1}^{n}deg(i)x_i)$, where $w_1$ and $rw_1=w_2$ (following Lemma~\ref{thm:weight_aligment}) are the learned parameters.

We now proceed to show that extrapolation to $P2$ fails in this case.
Let $G$ be a star graph, with features of dimension $1$ drawn from $\mathcal{N}(0,1)$, and assume w.l.o.g. that the center node of the star has index $1$.
Then applying the $f$ (learned on $P1$) to this $G$ can be written as $$f(G) = sign(w_1\sum_{i=1}^{n}\x_i+rw_1(n-1)x_1 + rw_1\sum_{i=2}^{n}x_i)$$

We will first show that when the number of vertices grows to infinity, the sign is determined by the first (central) node.
Denote
\begin{equation*}
\begin{array}{cccc}
     X = rw_1(n-1)x_1&  Y =  w_1\sum_{i=1}^{n}x_i & Z = rw_1\sum_{i=2}^{n}x_i & W=X+Y+Z
\end{array}
\end{equation*}

We will show that the correlation coefficient between $X$ and $W$ goes to $1$ as the number of vertices $n$ approaches infinity.
It holds that 
\begin{equation*}
\begin{array}{ccc}
X \sim \mathcal{N}(0, r^2w_1^2(n-1)^2) & Y\sim 
\mathcal{N}(0, w_1^2{n})
& Z\sim \mathcal{N}(0, r^2w_1^2(n-1))
\end{array}
\end{equation*}

\begin{equation*}
\begin{aligned}
    &Var(W) = Var(X+Z+Y) = Var(X+Z) + Var(Y) + 2COV(X+Z, Y) \\
    & COV(X+Z, Y) = COV(X,Y) + COV(Z,Y)\\
    & COV(X,Y) = COV(rw_1(n-1)x_1, w_1\sum_{i=1}^{n}x_i) = rw_1^2(n-1)COV(x_1,x_1) = rw_1^2(n-1)\\
    & COV(Y,Z) = COV(w_1\sum_{i=1}^{n}x_i, rw_1\sum_{i=2}^{n}x_i) = rw_1^2)\sum_{i=2}^{n}COV(x_i,x_i) = rw_1^2(n-1) \\
    & \implies COV(X+Z, Y) = 2rw_1^2(n-1) \\
    & \implies Var(W) = r^2w_1^2(n-1)^2 + r^2w_1^2(n-1) + w_1^2n + 2rw_1^2(n-1)\\
    & COV(X, W) = COV(rw_1(n-1)x_1, w_1\sum_{i=1}^{n}x_i+ rw_1\sum_{i=2}^{n}x_i+rw_1(n-1)x_1) \\
    & = rw_1(n-1) [ w_1COV(x_1, x_1) + rw_1(n-1)COV(x_1,x_1)] = rw_1^2(n-1) + r^2w_1^2(n-1)^2\\
    & \rho(X, W) = \frac{rw_1^2(n-1) + r^2w_1^2(n-1)^2}{(rw_1(n-1))(w_1\sqrt{r^2(n-1)^2 + r^2(n-1) + n + 2r(n-1)})} \\
    & =
\frac{rw_1^2(n-1)(1 + r(n-1))}{\sqrt{r^2(n-1)^2 + r^2(n-1) + n + 2r(n-1)})} = 
\frac{1 + r(n-1)}{\sqrt{r^2(n-1)^2 + r^2(n-1) + n + 2r(n-1)}}\\
& \frac{rn+1-r}{\sqrt{rn^2+(1+2r-3r^2)n-2r}} \rightarrow_{n \to \infty} 1 \\
\end{aligned}
\end{equation*}

As $X$ and $W$ are fully correlated, they have the same sign. 
We will now show that when $n$ approaches infinity, the probability that $X$ will have a different sign from $w_1^*\sum_{i=1}^{n}x_i$ is $0.5$, and therefore conclude that the error on $P2$ is $>0.25$ as specified in the theorem.
We will do so by showing that the correlation coefficient between $X$ and $w_1^*\sum_{i=1}^{n}x_i$ converges to $0$.

\begin{equation*}
\begin{aligned}
    COV(X, w_1^*\sum_{i=1}^{n}x_i) = r^w_1(n-1)w_1^*\\
    \rho(X, w_1^*\sum_{i=1}^{n}x_i) = \frac{rw_1(n-1)w_1^*}{rw_1(n-1) w_1^*\sqrt{n}} \rightarrow_{n \to \infty} 0
\end{aligned}
\end{equation*}

We conclude that a model trained on $P1$ will fail to extrapolate to $P2$.



\qedsymbol
\subsection{Proof of Theorem 3.3}
We consider the case of a graph-less teacher as in \eqref{eq:graphless_func} We wish to show that if the training data consists of infinitely many samples from a distribution over r-regular graphs, then the learned model will extrapolate perfectly to a distribution over r'-regular graphs. We assume the same feature distribution in all cases.
\comment{\amirg{this is not a new assumption. You can't make assumptions in the middle of the proof. The theorem states that the training data is separated by the learned model. Or it should say if it doesn't.}}

\comment{\amirg{Again this needs to be mapped to the thing you want to prove. Explain why what you do proves it.}\amirg{Say something like: We consider the case of a graph-less teacher as in \eqref{eq:graphless_func} We wish to show that if the training data consists of infinitely many samples from a distribution over r-regular graphs, then the learned model will extrapolate perfectly to a distribution over r'-regular graphs. We assume the same feature distribution in all cases.}}

\comment{
Let $S$ be a graph dataset with features drawn from a distribution $D$ and graphs drawn from an $r$-regular graph distribution $D_r$.
Assume that the label generator function of $S$, denoted by  $f^*$, is generated by a teacher GNN with $\w_2^*=0$, i.e., for all $G\in S, 1\leq l\leq m$, $f^*(G) =\w_1^*\sum_{i=1}^n x_i$. 
\comment{as follows from Equation~\ref{eq:max_margin_regular}.}
}

Let $f=[\w_1, \w_2]$ be a minimizer of Equation~\ref{eq:max_margin_regular} on the training distribution. Then $f$ has perfect accuracy on the support of the training distribution (ie it is equal to the graph-less teacher $f^*$ there).
Let $G=(X_G, A_r)$ be in the support of the training distribution. Then:
\begin{equation}
\begin{aligned}
&f(G) = sign\left((\w_1 +r\w_2)\tilde{x}_G\right) \stackrel{(*)}{=} sign\left((\w_1 +r^2\w_1)\tilde{x}_G\right)
\stackrel{(**)}{=}sign\left(\w_1\tilde{x}_G\right) \stackrel{(***)}{=}sign\left(\w_1^*\tilde{x}_G\right)
\end{aligned}
\end{equation}
(*) Follows from Theorem 3.1 by substituting $\w_2 = r\w_1$.
(**) Follows from the fact that the direction of $(\w_1 +r^2\w_1)$ and $\w_1$ is the same. (***) Follows from the fact that $f$ is equal to $f^*$ on the training distribution.
\comment{\amirg{you didn't specify this assumption. Need to be clearer here about it.}}

Now let $G_{r'}=(X_{G_{r'}}, A_{r'})$ be an $r'$-regular graph example, with features drawn from $D$.
Following Equation 2, we get that:
\begin{equation*}
\begin{aligned}
&f(G_{r'})=sign\left((\w_1 +r'\w_2) \tilde{x}_{G_{r'}} \right) = sign\left((\w_1 +r'r\w_1) \tilde{x}_{G_{r'}} \right) 
\stackrel{(***)}{=} sign\left(\w_1 \tilde{x}_{G_{r'}} \right) = sign\left(\w^*_1 \tilde{x}_{G_{r'}} \right)
\end{aligned}
\end{equation*}

(***) Follows from the assumption that the features are drawn from $D$.
We thus have that all instances drawn from the test distribution of r'-regular graphs are classified correctly, and therefore we have perfect extrapolation in this case.
\qedsymbol
\comment{
\subsection{Proof of Lemma 3.4}

We saw that a linear GNN trained on a distribution with regular graphs of degree $r$ and labeled via a linear function $w_1^*$ generalizes to any $r'$. Here we show that there exist ``bad solutions'' that solve the in-distribution problem (i.e r-regular graphs) but do not generalize to r'-regular graphs. We note that if one does not train with GD, one may learn these bad solutions, and generalize poorly. This highlights the importance of training with GD because of its implicit bias in this case.\\

Let $S=\{(X^{(l)},A^{(l)},y^{(l)}_l)\}$ be a graph dataset labeled by a graph-less teacher of
\comment{a linearly separable\amirg{I dont' think you should say linearly separable. You should say it is labeled by a graph-less teacher.}} 
$r$-regular graph dataset with $X^{(n)} \in \mathbb{R}^{n \times d}$ being the node features and $y_l=sign(w^*\sum_{i=1}^n x_i)$.
We will now show that there exists a GNN with parameters $u_1, u_2 \in \mathbb{R}^d$ that fits $S$ perfectly and fails to generalize for any graph with regularity degree $r'\neq r$.

Let $w_g$ be some classifier\comment{\amirg{not clear? Where is this coming from? Is this the graph-less teacher?}} with unit margin obtained on S, i.e., $y_l \langle w_g, \sum_{i=1}^n x_i \rangle \geq 0 \quad \forall n$ and $y_l \langle w_g, \sum_{i=1}^n x_i \rangle \geq 1$. 
\comment{\amirg{why not say $\geq 1$? Why do you need both?}}
Set $w_b = \frac{-1.5y_s}{||\sum_{i=1}^n x_i^{(s)}||^2} \sum_{i=1}^n x_i^{(s)}$ which implies that: $y_s \langle w_b, \sum_{i=1}^n x_i^{(s)} \rangle = -1.5$\\

Let $G$ be an $r'$-regular graph, with $r' >r$.
Let $u_1 = w_g - r w_b, \quad u_2 = w_b$ then $u_1 + r u_2 = w_g$.
Then a GNN with parameters $u_1$ and $u_2$ fits $S$ with accuracy $1$, but

$$ y_s \langle u_1 + r'u_2, \sum_{i=1}^n x_i^{(s)} \rangle  = y_s \langle w_g + (r'-r)w_b, \sum_{i=1}^n x_i^{(s)} \rangle = $$
$$1 + (r'-r)(-1.5) \leq -0.5$$
Therefore the above GNN will have an error of $1$ on all r'-regular graphs, and fail to extrapolate.

The result for $r'<r$ can be shown similarly.
The result also applies when $n\to\infty$
\qedsymbol
}
\subsection{Proof of Theorem 3.4}

Let $f^*$ be a graph-less function as in \eqref{eq:graphless_func}, and 
 $f$ be a GNN minimizing Equation~\ref{eq:max_margin_regular}, on a training set of $r$-regular graph examples. Assume we have modified an example in $S$ from $G=(X,A)$ to $\tilde{G} = (X,\tilde{A})$.
 Let $\tilde{x} = \sum_{i=1}^{n}\x_i,~ \x_i\in G$. Let $0\leq r' \leq n-1, ~ r'\in \mathbb{N}$ and let $\Delta_{r',\tilde{G}}(i) = deg_{\tilde{G}}(i)-r'$.\\
Then using Equation~\ref{eq:max_margin_general}:
\begin{equation}
    \begin{aligned}
    &f(\tilde{G}) =\w_1\tilde{x} + \w_2\sum_{i=1}^{n}deg(i)\x_i \stackrel{(*)}{=} \w_1\tilde{x} + r\w_1\sum_{i=1}^{n}deg(i)\x_i \\
    & = \w_1\tilde{x} + r\w_1\sum_{i=1}^{n}(\Delta_{r',\tilde{G}}(i) + r')\x_i \\
    &= \w_1\tilde{x} + r\w_1\sum_{i=1}^{n}\Delta_{r',\tilde{G}}(i) + r'r\w_1\sum_{i=1}^{n}\x_i \\
    & =  \w_1\tilde{x} + r'r\w_1 \tilde{x} + r\w_1\sum_{i=1}^{n}\Delta_{r',\tilde{G}}(i)\\
    & = \underbrace{\w_1\tilde{x} + r'r\w_1\tilde{x}}_{Regular~Component} + \underbrace{r\w_1\sum_{i=1}^{n}\Delta_{r',\tilde{G}}(i)\x_i}_{\Delta~Component}
    \end{aligned}
\end{equation}

Where (*) follows from Theorem 3.1.

Now assume there exists an $r'$ such that:
\begin{equation*}{\label{eq:extrapolation_ratio}}
\mid\frac{r\w_1\sum_{i=1}^{n}\Delta_{r',\tilde{G}}(i)\x_i}{\w_1\tilde{x} + r'r\w_1\tilde{x}}\mid \leq 1 
\end{equation*}

Therefore the $\Delta$-component is small with respect to the regular component, and can be dropped below, because it doesn't change the sign.
$$
f(\tilde{G}) = sign(\w_1\tilde{x} + r'r\w_1\tilde{x}) \stackrel{(**)}{=} f^*(\tilde{G})
$$
Where $(**)$ follows from Theorem 3.3.

\section{Additional Experimental Results}

\subsection{Empirical Validation of Lemma 3.1}
\begin{figure}[!h]
    \centering
    \includegraphics[width=0.7\textwidth]{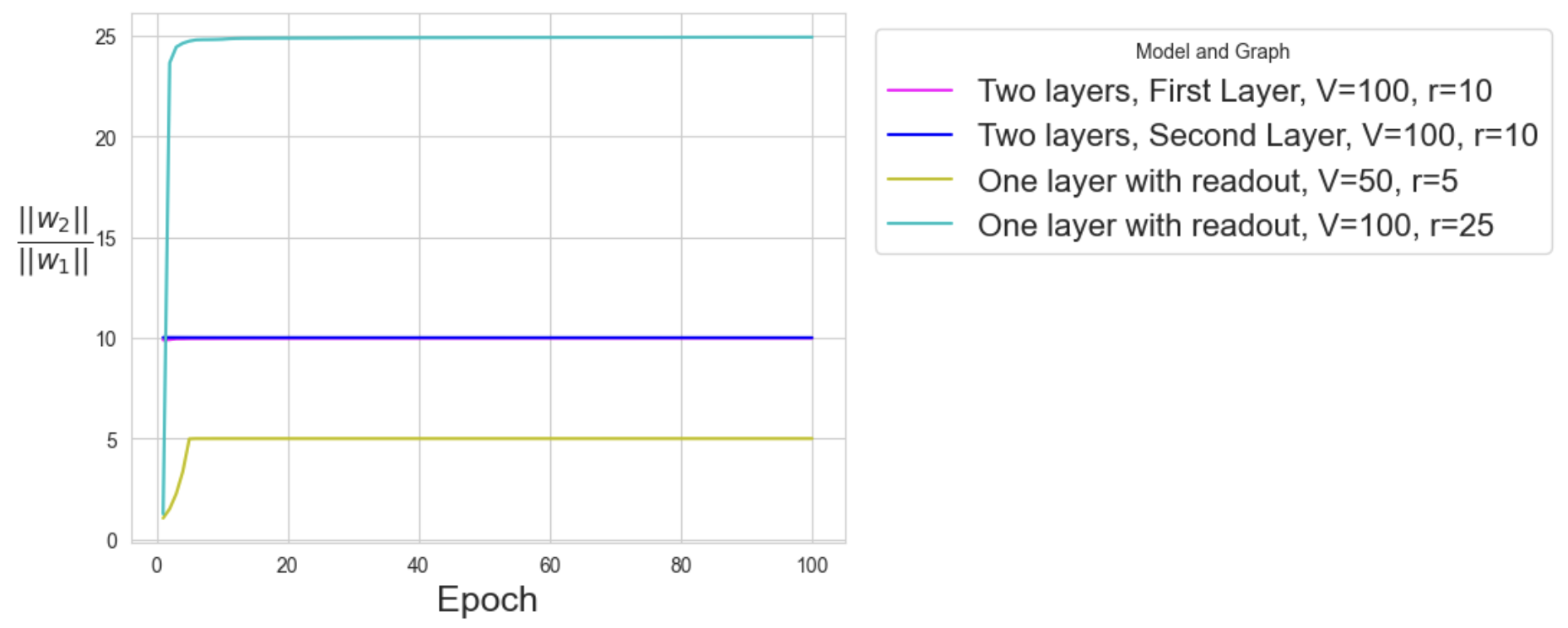}
    \caption{An empirical validation of Theorem 3.1. The ratio between the topological and root weights is equal to the regularity degree of the graphs. V is the number of nodes in each graph, and r is the regularity degree. }
    \label{Figure:raguler_ratio}
\end{figure}

The validation of Theorem 3.1 is presented in  Figure~\ref{Figure:raguler_ratio}. We plot the ratio between the topological weights and root weights, during the training of linear GNNs with one or two layers, with readout. The GNNs are trained on regular graphs with different regularity degrees. In all cases, the ratio converges to the regularity degree, as guaranteed by Theorem 3.1.

\subsection{Evidences For Graph Overfitting With Additional Models (Section 2.2)}
\begin{figure*}[h]
    \centering
    \includegraphics[width=0.7\linewidth]{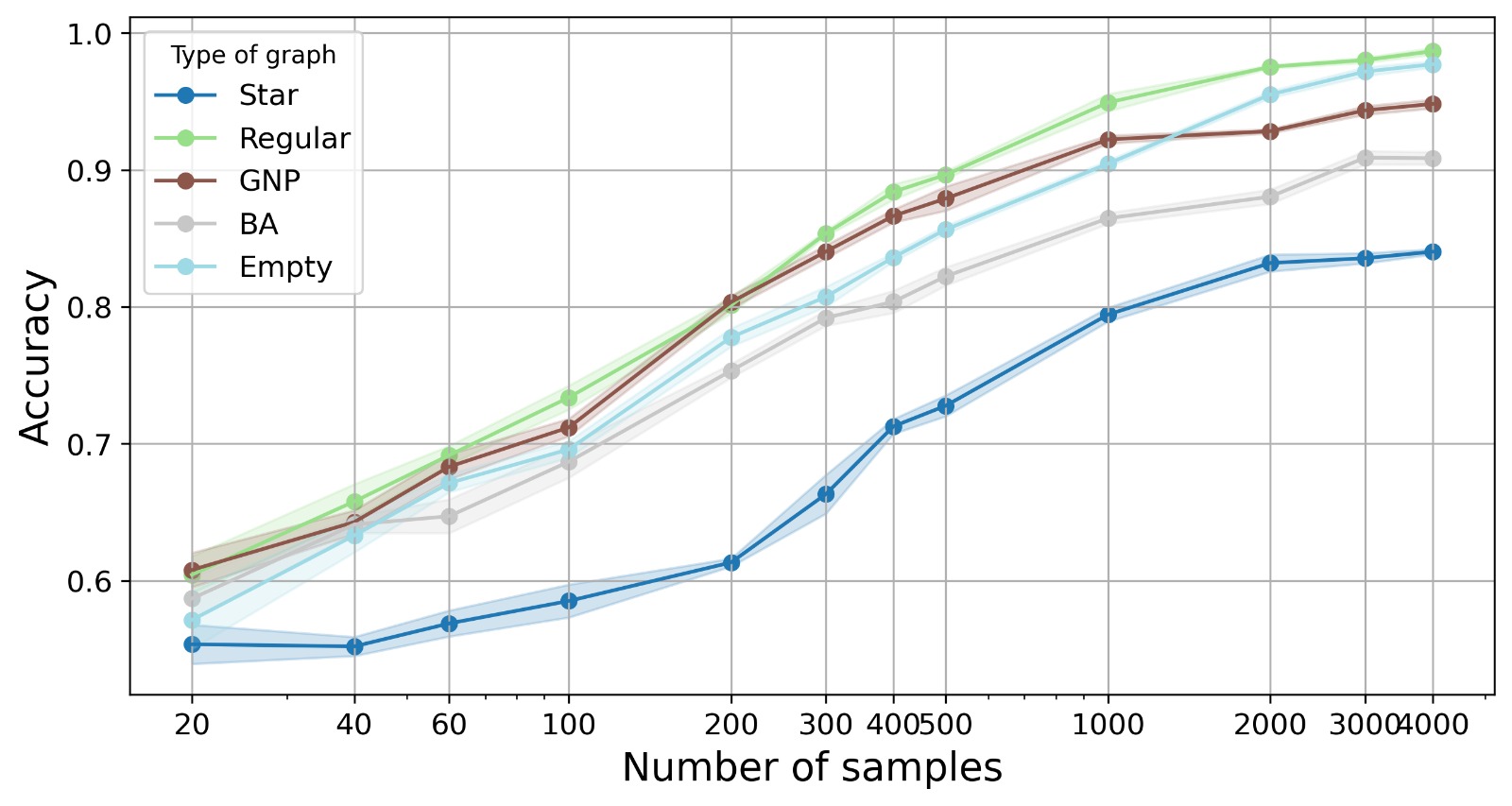}
    \caption{Evaluation of the GIN~\citep{gin} model on the Sum task where the graph should be ignored, as described in Section 2.2 in the main paper.}
    \label{Figure:gin_plot}
\end{figure*}

\begin{figure*}[h]
    \centering
    \includegraphics[width=0.7\linewidth]{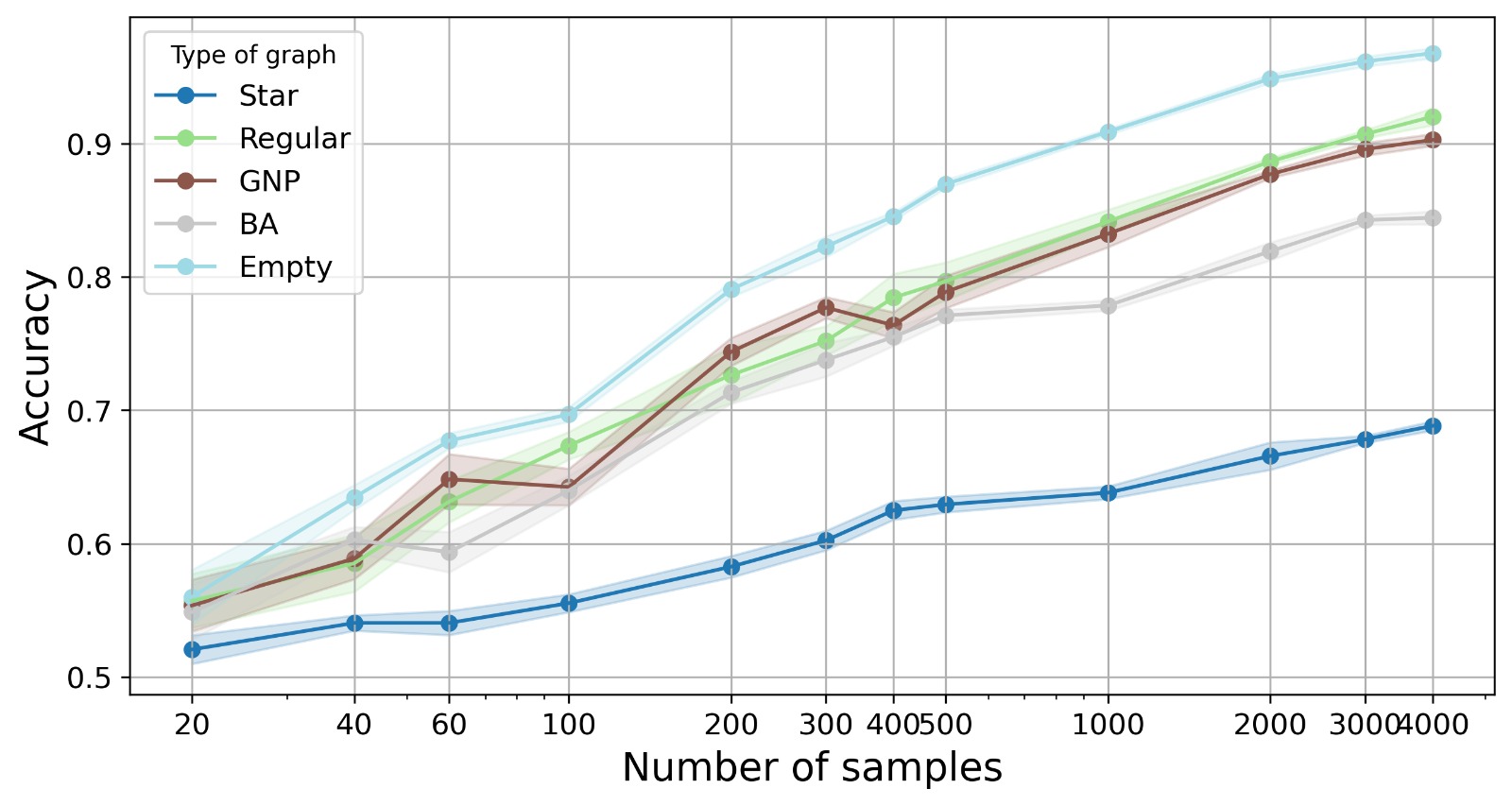}
    \caption{Evaluation of the GAT~\citep{gat} model on the Sum task where the graph should be ignored, as described in Section 2.2 in the main paper.}
    \label{Figure:gat_plot}
\end{figure*}

\begin{figure*}[h]
    \centering
    \includegraphics[width=0.7\linewidth]{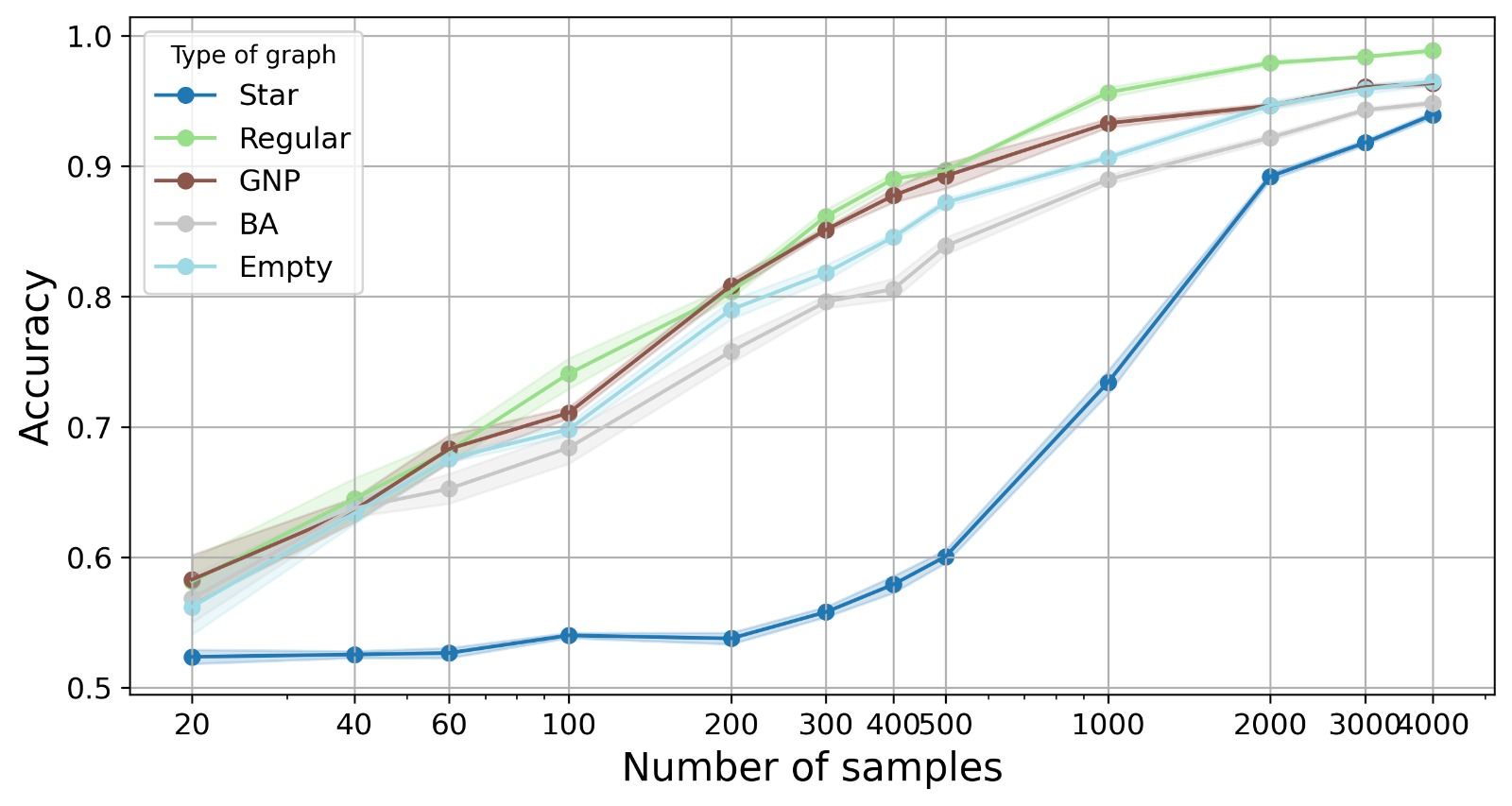}
    \caption{Evaluation of the Graph Transformer~\citep{shi2021masked} model on the Sum task where the graph should be ignored, as described in Section 2.2 in the main paper.}
    \label{Figure:transormer_plot}
\end{figure*}

\begin{figure*}[h]
    \centering
    \includegraphics[width=0.7\linewidth]{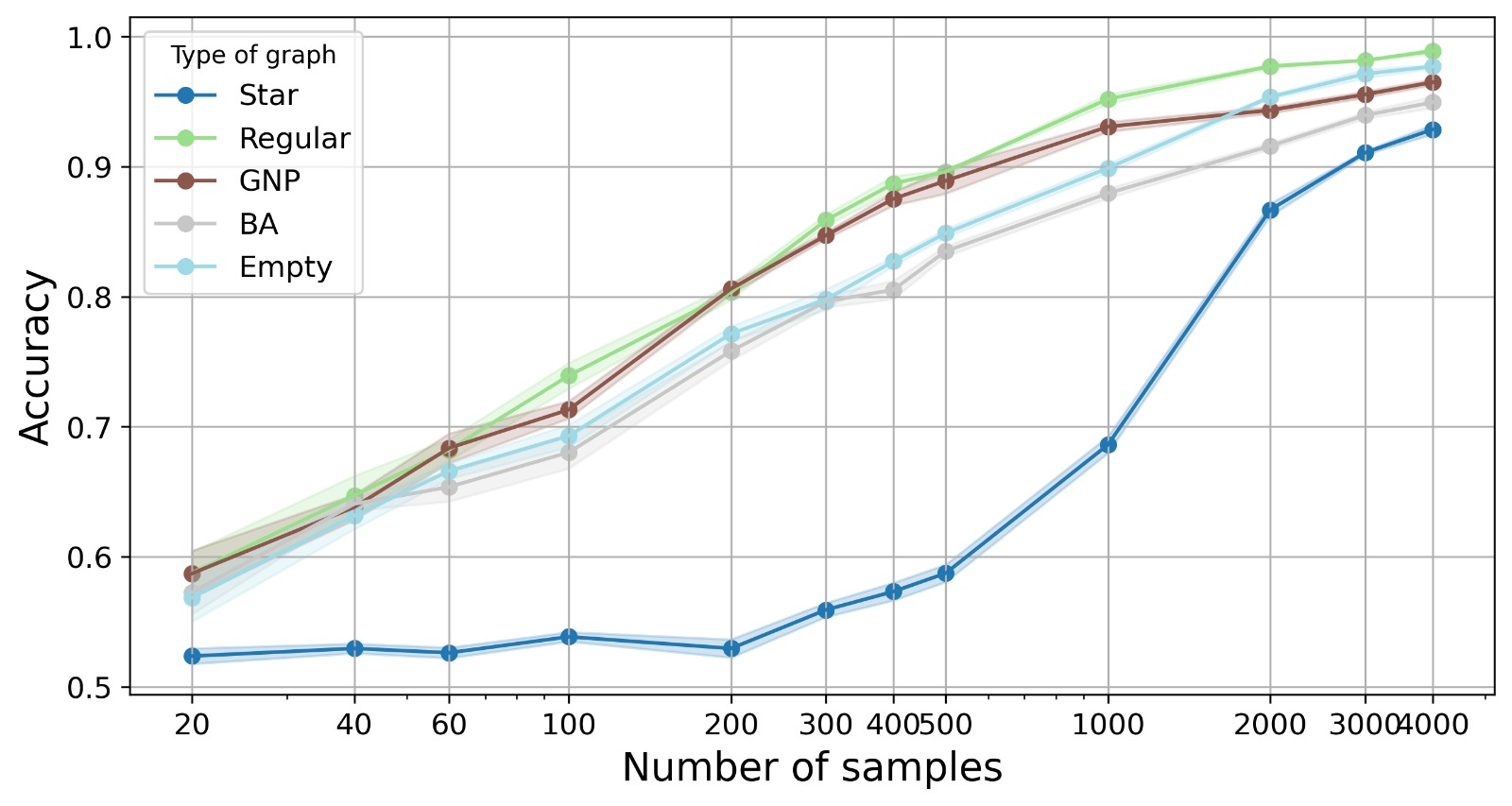}
    \caption{Evaluation of the same model presented in Equation~\ref{eq:gnn}~\citep{mpgnn} with normalized neighbor aggregation on the Sum task where the graph should be ignored, as described in Section 2.2 in the main paper. }
    \label{Figure:mean_agg_plot}
\end{figure*}

 In Section 2.2 we presented an empirical evaluation of the model from \citet{mpgnn} as described in Equation~\ref{eq:gnn}. Here we provide the results of the same evaluation with more GNNs. All models show similar trends as presented in the main paper.
 The results are shown in Figures \ref{Figure:gin_plot} 
 (GIN~\citep{gin}), \ref{Figure:gat_plot} (GAT~\citep{gat}), \ref{Figure:transormer_plot} (Graph Transformer~\citep{shi2021masked}) and \ref{Figure:mean_agg_plot} (GraphConv with Normalized Neighbor Aggregation~\citep{mpgnn}).

\subsection{Evidence For Graph Overfitting With Node Task (Section 2.2)}
\begin{figure*}[h]
    \centering
    \includegraphics[width=0.6\linewidth]{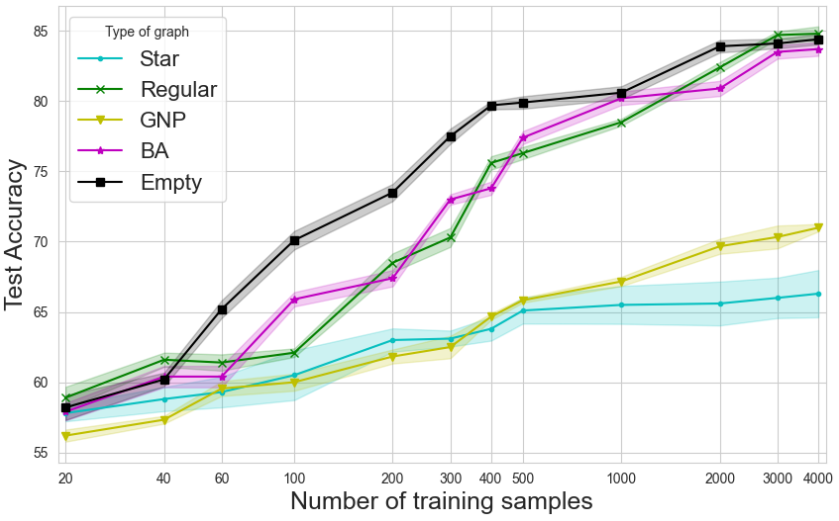}
    \caption{Evaluation of the same model presented in Equation~\ref{eq:gnn}~\cite{mpgnn} on the Sum task for node classification where the graph should be ignored.}
    \label{Figure:node_task}
\end{figure*}

We evaluated the learning curve in a teacher-student setup of a graph classification task, where the teacher is graph-less GNN. 
The teacher readout is sampled once from $\mathcal{N}(0,1)$ to generate the train, validation and test labels. The training graph is over $4000$ nodes and the validation and test graphs are over $500$ nodes. Each node is assigned with a feature vector in $\mathbb{R}^{128}$ sampled i.i.d from $\mathcal{N}(0, 1)$.
Figure~\ref{Figure:node_task} shows that also in this case, although the teacher does not use the graph, giving the model different graphs affects generalization. Therefore also in this case, the GNN overfits the given graph although it should be ignored. 

\subsection{Additional Results on How Graph Structure Affects Overfitting (Section 2.3)}
\begin{figure*}[t]
    \centering
    \includegraphics[width=0.7\linewidth]{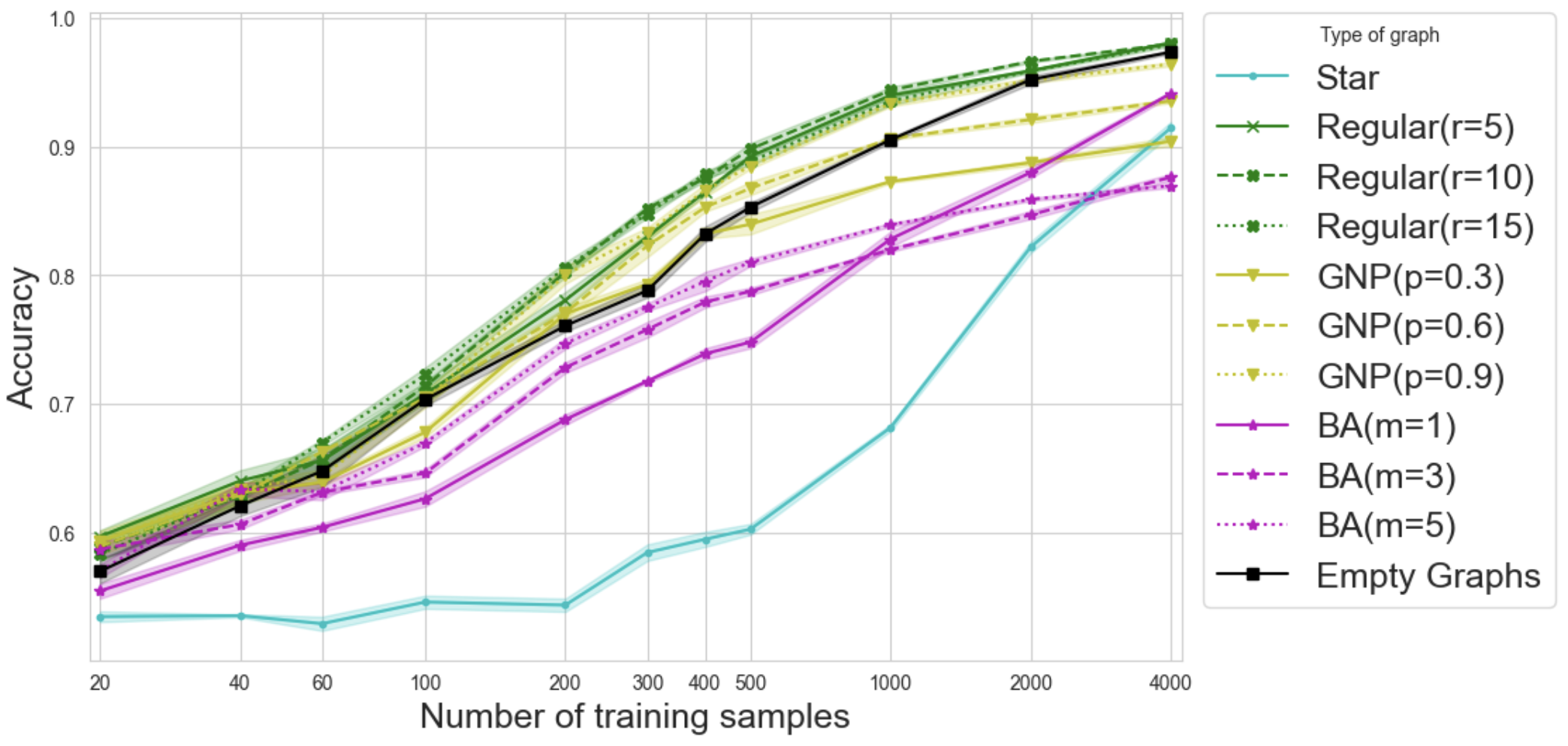}
    \caption{The learning curves of the same GNN model trained on graphs that have the same node features and only differ in
their graph-structure. The label is computed via a graphless teacher. If GNNs were to ignore the non-informative graph-structure they were given, similar performance should
have been observed for all graph distributions. Among the different distributions, regular graphs exhibit the best performance.}
    \label{Figure:sample_complexity_full}
\end{figure*}

\begin{figure*}[ht]
    \centering
    \includegraphics[width=0.7\linewidth]{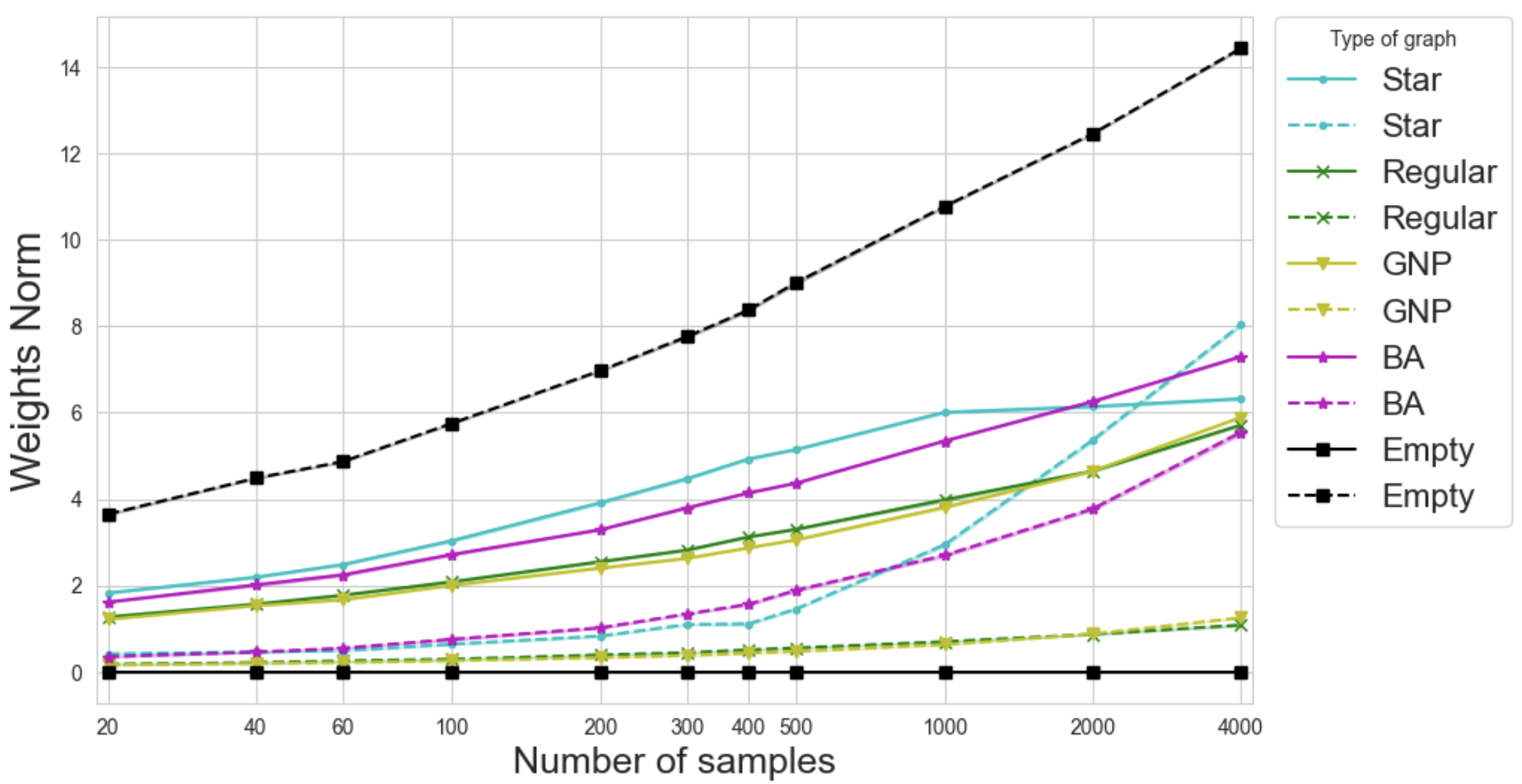}
    \caption{The weights norm of the topological (dashed) and the root (smooth) weights along the same runs. On the empty graphs, the topological
weights are not trained and the ratio is 0 due to initialization.}
    \label{Figure:weight_norm}
\end{figure*}

In Section 2.3 for the sake of presentation, we presented only one curve from each distribution. Figure~\ref{Figure:sample_complexity_full} presents the learning curve of all the distributions we tested, with multiple parameters for each distribution.
Additionally, in Figure~\ref{Figure:weight_norm} we present the weights norms of the root and topological weights separately, for the curves presented in the main paper.

\section{Additional Experimental Details}
\subsection{Additional Implementation Details For Section 2.2}

The teacher readout is sampled once from $\mathcal{N}(0,1)$ and used for all the graphs. 
All graphs have $n=20$ nodes, and each node is assigned with a feature vector in $\mathbb{R}^{128}$ sampled i.i.d from $\mathcal{N}(0, 1)$.

 For the Sum task, we used a $1$-layer ``student" GNN following the teacher model, with readout and ReLU activations. For the PROTEINS and ENZYMES tasks, we used $3$-layers.

 We evaluated the learning curve with an increasing amount of $[20, 40, 60, 100, 200, 300, 400, 500, 1000, 2000, 4000]$ samples.
We note that the GNN has a total of $\sim$16,000 parameters, and thus it is overparameterized and can fit the training data with perfect accuracy.

\subsection{Experimental setup in Section~\ref{sec:extrapolation_sufficient_condition}}

We trained a one-layer linear GNN with readout on $5$-regular graphs over $20$ nodes.
We then applied it to a test sets presented in Table~\ref{table:regular_extrapolation}. Each test set contains $100$ graph examples and each graph has $20$. All the test sets share the same node features and differ in the graph structure, which is drawn from different graph distributions.

\subsection{Additional Dataset Information}

\begin{table*}[!h]
  \centering 
  \caption{Statistics of the real-world datasets used in our evaluation.}
  \label{Table:data_sum}
\begin{tabular}{lccccc} 
\\
Dataset  & \# Graphs & Avg \# Nodes & Avg \# Edges & \# Node Features &\# Classes  \\ 
   \hline\\

Proteins   & 1113     & 39.06       & 72.82    &3   & 2          \\ 
NCI1       & 4110     & 29.87       & 32.3     &37   & 2         \\ 
Enzymes        & 600      & 32.63          & 62.14      &3    & 6          \\
D\& D  & 1178     & 284.32      & 715.66  &89   & 2          \\ 
IMDB-B    & 1000     & 19          & 96     &0     & 2          \\ 
IMDB-M    & 1500     & 13          & 65    &0      & 3          \\ 
Collab & 5000 & 74.49 & 2457.78 & 0 & 3\\
Reddit-B & 2000 &	429.63 &	497.75 & 0& 2\\
Reddit-5k & 4999 & 508.52 & 594.87 & 0 & 5\\
mol-hiv & 41,127	& 25.5 & 27.5 & 9&2\\
mol-pcba & 437,929&	26.0	&28.1&	9 & 2 (128 tasks) \\

\bottomrule
\end{tabular}

\end{table*}

The dataset statistics are summarized in Table~\ref{Table:data_sum}.

\textbf{IMDB-B \& IMDB-M}~\cite{dgk} are movie collaboration datasets. Each graph is derived from a genre, and the task is to predict this genre from the graph. Nodes represent actors/actresses and edges connect them if they have appeared in the same movie.

\textbf{Proteins, D\&D \&Enzymes}~\cite{wl, proteins} are datasets of chemical compounds. The goal in the first two datasets is to predict whether a compound is an enzyme or not, and the goal in the last datasets is to classify the type of an enzyme among $6$ classes.

\textbf{NCI1}~\cite{wl} is a datasets of chemical compounds. Vertices and edges represent atoms and the chemical bonds between them. The graphs are divided into two classes according to their ability to suppress or inhibit tumor growth.

\textbf{Collab}~\cite{tudataset} is a scientific collaboration dataset. A graph corresponds to a researcher’s ego network, i.e., the researcher and their collaborators are nodes and an edge indicates collaboration between two researchers. A researcher’s ego network has three possible labels, which are the fields that the researcher belongs to. 

\textbf{Reddit-B, Reddit-5k}~\cite{tudataset} are datasets of Reddit posts from the month of September 2014, with binary and multiclass labels, respectively. The node label is the community, or ``subreddit", that a post belongs to. $50$ large communities have been sampled to build a post-to-post graph, connecting posts if the same user comments on both.

\textbf{mol-hiv, mol-pcba}~\cite{ogb} are large-scale datasets of molecular property prediction.

 Following~\citet{errica2022fair}, we added a feature of the node degrees for datasets which have no node features at all.
\subsection{Hyper-Parameters}
 All GNNs use ReLU activations with $\{3,5\}$ layers and $64$ hidden channels. They were trained with Adam optimizer over $1000$ epochs and early on the validation loss with a patient of $100$ steps, eight Decay of $1e-4$, learning rate in $\{1e-3, 1e-4$\}, dropout rate in $\{0, 0.5\}$, and a train batch size of $32$.
The preserved COV is among \{80\%, 50\%\}.
All the experiments' code including the random seeds generator is provided in the code Appendix


\subsection{Models}
In Section~\ref{sec:experiments} when evaluating graphs with reduced COV, we add edge features to differ between the original and added edges. We adapt each neighbor's aggregation component to process this edge information in a non-linear way. 

\paragraph{GraphConv}

$$\x' = W_1\x_i + W_2\sum_{j\in N(i)} x_i + \sigma(W_3\e_{i,j})$$

\paragraph{GIN}

$$\x' = W_4\left((1+\epsilon)\x_i + W_2\sum_{j\in N(i)} \sigma\left(\x_i + \sigma(W_3\e_{i,j})\right)\right)$$

\paragraph{GATv2}

$$x' = \alpha_{i,i}W_1\x_i + \sum_{j\in N(i)} \alpha_{i,j}W_2\x_j$$

$$
\alpha_{i,j} = \frac{exp\left(a^TLeakyReLU\left(W_1\x_i + W_2\x_j + \sigma(W_3\e_{i,j})\right)\right)}{\sum_{k\in N(i)\cup \{i\}}exp\left(a^TLeakyReLU\left(W_1\x_i + W_2\x_k + \sigma(W_3\e_{i,k})\right)\right)}
$$

\paragraph{GraphTransformer}

$$x' = W_1\x_i + \sum_{j\in N(i)} \alpha_{i,j}\left(W_2\x_j + \sigma(W_5\e_{i,j})\right)$$

$$
\alpha_{i,j} = \frac{(W_3\x_i)^T(W_4\x_j+W_5\e_{i,j})}{\sqrt{d}}
$$



\end{document}